\documentclass{article}

\usepackage[final]{corl_2019} % Uncomment for the camera-ready ``final'' version

\usepackage{amsmath}
\usepackage{amsthm}
\usepackage{amssymb}
\usepackage{color}
\usepackage{algorithm}
\usepackage[noend]{algpseudocode}
\usepackage{array}
\usepackage{graphicx,subcaption}
\usepackage[page]{appendix}
\usepackage{dsfont}
\usepackage{wrapfig,lipsum,booktabs}

\title{Combining Deep Learning and Verification for Precise Object Instance Detection}

\author{
  Siddharth Ancha\thanks{Equal contribution},~ Junyu Nan\footnotemark[1],~ David Held\\
  Carnegie Mellon University\\
  United States\\
  \texttt{\{sancha, jnan1, dheld\}@andrew.cmu.edu} \\
}

\newcommand{\flowverify}[0]{\textsc{FlowVerify}}
\newcommand{\siftverify}[0]{\textsc{SIFTVerify}}
\newcommand{\flowmatchnet}[0]{\textsc{FlowMatchNet}}

\newcommand{\fp}[0]{\textsc{FPrecision}}
\newcommand{\fr}[0]{\textsc{FRecall}}
\newcommand{\finlier}[0]{\textsc{FRigidity}}

\newcommand{\nms}[0]{\textsc{SimObj}}
\newcommand{\fcolor}[0]{\textsc{FColor}}

\begin{document}
\maketitle

%===============================================================================

\begin{abstract}
    Deep learning object detectors often return false positives with very high confidence.
    % Object detectors are typically trained using machine learning algorithms on lots of data.
    Although they optimize generic detection performance, such as mean average precision (mAP), they are not designed for \textit{reliability}.
    For a reliable detection system, if a high confidence detection is made, we would want high certainty that the object has indeed been detected.
    To achieve this, we have developed a set of verification tests which a proposed detection must pass to be accepted. We develop a theoretical framework which proves that, under certain assumptions, our verification tests will not accept any false positives. Based on an approximation to this framework, we present a practical detection system that can verify, with \textit{high precision}, whether each detection of a machine-learning based object detector is correct.
    We show that these tests can improve the overall accuracy of a base detector and that accepted examples are highly likely to be correct. This allows the detector to operate in a high precision regime and can thus be used for robotic perception systems as a reliable instance detection method.
    Code is available at \href{https://github.com/siddancha/FlowVerify}{\texttt{https://github.com/siddancha/FlowVerify}}.
\end{abstract}

%Our verification checks are based on estimated dense pixel correspondences between known images of objects and a scene; however, our theoretical framework does not make assumptions about the .


%Our method is \textit{one-shot}, in that works for novel objects never seen during training, and our model can be used off-the-shelf without any retraining or finetuning, for reliable instance detection. 

%===============================================================================

\section{Introduction}

Instance detection is the task of detecting instances of a particular object in a scene. Here (as in previous work~\cite{xie2013multimodal,held2016robust,georgakis2016multiview}), the term ``instance" refers to a specific sub-type of object (e.g. ``coke can" rather than just ``can"). For example, a robot may be shown an example image of a cereal box, and it may be required to detect it in a scene in order to fetch it, even if it is in a slightly different viewpoint than the example image.
% There are many real-world scenarios in which we want to detect objects in a scene from a pre-defined set of object instances. Here (as in previous work~\cite{Singhetal_ICRA2014,xie2013multimodal,held2016robust,georgakis2016multiview}), the term ``instance" refers to a specific sub-type of object (e.g. ``coke can" rather than just ``can").  For example, many product databases available on Amazon or other product websites contain household products that are likely to be found in a person's home.  A smart home perception system would benefit from being able to reliably detect objects from this database.  Similarly, a kitchen perception system would benefit from being able to match  grocery products in the kitchen to a database obtained from the local grocery store.
% Additionally, for industrial robots or other applications where a robot is placed in a fixed environment, a robot might be need to interact with only a pre-defined set of objects.
A user may wish to give a robot instructions that refer to a specific object, e.g. to bring the user's coffee mug (as opposed to a random coffee mug).
Unfortunately, current systems for object instance detection are not sufficiently reliable for use in real world applications; most methods will fail in real-world scenarios due to occlusions, lighting changes, viewpoint variation, and other difficulties.   

% \begin{figure}[H]
% 	\centering
% 	\includegraphics[width=0.7\textwidth]{pics/home2.pdf}
% 	\caption{Left: Clean product images from a product database; Right: products found in a  home environment.  There are many challenges of reliably detecting products in such a cluttered setting.}
% 	\label{fig:pull}
% \end{figure}

% Traditional non-parametric methods for object instance detection rely on keypoint-matching~\cite{lowe2004distinctive, bay2006surf,
% 	quadros2012occlusion} or template matching~\cite{huttenlocher1993comparing,hinterstoisser2012gradient} to training images. However, these methods are not robust to large changes in object viewpoint (i.e. greater than 25 degrees~\cite{held2016robust}) and often fail for significant lighting changes.  For a robust home perception system, we cannot always guarantee that the objects being observed will be viewed from the same conditions as in training.  
	
%	Such methods are often combined with geometric verification techniques such as RANSAC~\cite{fischler1981random}.  

\begin{figure}
  \centering
  \includegraphics[width=0.74\textwidth]{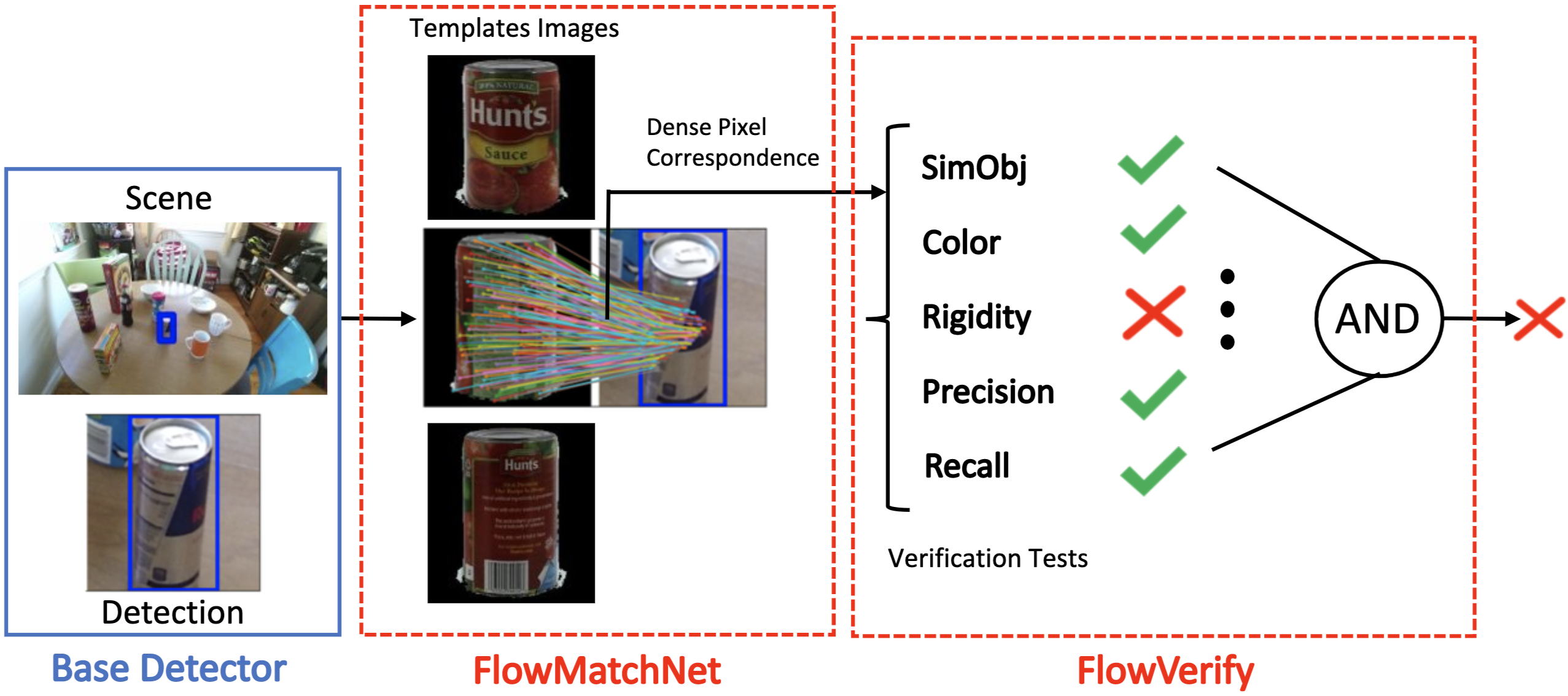}
  \caption{Pipeline of our instance detection verification system. \textit{Base Detector}: generates proposed object instance detections; \textit{\flowmatchnet}: computes dense pixelwise correspondences between template images of an object and a proposed detection; \textit{\flowverify}: a suite of verification tests is applied to the detection using the estimated correspondences. The detection is accepted only if all verification tests pass.}
  \label{fig:pipeline}
  \vspace{-5pt}
\end{figure}

Traditional non-parametric methods for object instance detection rely on keypoint-matching~\cite{lowe2004distinctive, bay2006surf,
	quadros2012occlusion} or template matching~\cite{huttenlocher1993comparing,hinterstoisser2012gradient} to a set of template images. However, these methods are not robust to large changes in object viewpoint (i.e. greater than 25 degrees~\cite{held2016robust}) and often fail for significant lighting changes.  For a robust home perception system, we cannot always guarantee that the objects being observed will be viewed from the same conditions as in the templates.  Thus we desire to have an object detection system that is robust to significant changes in the object viewpoint, as well as lighting, occlusions, and other variations. Recently, a number of machine learning approaches have been used for object instance detection~\cite{lai2011large,georgakis2017synthesizing,held2016robust}.  However, machine-learning based approaches often produce a large number of false positive detections.  These false detections can prevent deployment of robots in real-world applications.  
  
  % eventually discard training examples,
  %, that can also improve overall detection performance
  
  Our first insight is that a combination of parametric and non-parametric approaches can be used to obtain both robust and reliable object detection. While machine learning methods enable us to leverage large amounts of data to learn necessary invariances like lighting and viewpoint changes, non-parametric methods that match candidate detections to  template images can ensure that these detections are reliable. In sections~\ref{sec:theoretical-framework} and~\ref{subsec:verification-tests} we propose verification tests that operate on top of existing machine learning detectors; these tests take proposed detections as input and verify them by matching against template images. Thus, by not throwing away training data (in contrast to the parametric machine learning approach) and using them to verify detections at test time, we show that we can increase the accuracy of the detector, especially in the high-precision regime. This makes our system more precise, reliable and interpretable, making it more suitable for robotics applications.

  Our approach is based on a novel  theoretical framework for verified object detection. Specifically, under our theoretical framework, we prove that under certain conditions, a series of verification tests will reject all false positive detections. Under these assumptions, any detection that is verified by our tests is guaranteed to be correct. These verification tests are based on estimated dense correspondences between the proposed detection and a set of template images.  It is important to note that our theoretical framework does not make assumptions on the accuracy of the base detector or the accuracy of the estimated correspondences; the tests can fail due to either a false positive proposed by the detector or due to inaccurately estimated correspondences.  However, the tests will  pass only if the detection is a true positive.   
  
  Based on our theoretical framework, we implement an approximate but more practically suitable version of these tests.  In our practical framework, if any of the tests fail, rather than rejecting the detection, we reduce its initial confidence score.   We show that our method improves the performance of the detector in the high precision regime and performs no worse in the low-precision regime, thereby improving the overall accuracy of the detector. Figure \ref{fig:pipeline} shows a diagram of our method.

Because our system matches proposed detections to template images, the runtime is a function of the number of template images used.  However, the number of template images can be varied to obtain an accuracy-speed tradeoff; we demonstrate that the user of our system can use more templates from different viewpoints to achieve higher accuracy, or they can choose to use fewer templates to increase speed. We summarize our contributions as:
\begin{itemize}
    \item A theoretical framework for verified object instance detection, which guarantees, under certain assumptions, no false positive detections.
    \item An approximate implementation of this framework which leads to a practical instance detection method
    \item A demonstration that our method leads to significantly improved detector performance, especially in the high-precision regime.
    %\item A novel approach for object detection that combines parametric learned bounding box detection, parametric learned correspondence matching, and non-learned verification tests.
\end{itemize}
\vspace{-10pt}

\section{Related Work}
\label{sec:related-work}

%However, these methods are not robust to large changes in object viewpoint (i.e. greater than 25 degrees~\cite{held2016robust}) and often fail for significant lighting changes.  For a robust home perception system, we cannot always guarantee that the objects being observed will be viewed from the same conditions as in training.  
%become a standard for general object detection. Instance detection is slightly different because there is lesser intra-class variation and a few template images per object are allowed. 

\textbf{Object Instance Detection}. 
Traditional methods for object instance detection rely on keypoint-matching~\cite{lowe2004distinctive, bay2006surf,
	quadros2012occlusion} or template matching~\cite{huttenlocher1993comparing,hinterstoisser2012gradient}. Recently, detectors based on deep neural networks have shown improved performance over these traditional approaches.  For example, Target Driven Instance Detection (TDID)~\cite{tdid} is a state of the art instance detection method based on a siamese  neural architecture~\cite{koch2015siamese}.  We compare to TDID in our work and show improved performance over this state-of-the-art baseline.

%Instance detection is also related to navigation~\cite{zhu2017target} and tracking~\cite{held2016learning} since both of these applications involve matching objects in previous frames (templates) to the current frame.
 
\textbf{Grocery product recognition}. There has been significant effort in recognizing products on shelves of retail stores such as grocery product recognition~\cite{merler2007recognizing, george2014recognizing, franco2017grocery, geng2018fine}. This problem is simpler than the general object instance recognition problem that we are aiming to solve due to the structured environment.  For example, objects in such scenes are typically placed in front-facing canonical viewpoints with minimal occlusions and good lighting conditions.  Furthermore, there is often a great deal of contextual information, such as the location of the product on the shelf.  Our method is designed for the more general usage in the home or other unstructured locations.

\textbf{Dense Pixel Correspondence.} 
Our work builds on past work for computing dense pixel correspondences~\cite{choy2016universal,florence2018dense,dosovitskiy2015flownet,ilg2017flownet}; unlike past work, we use these correspondences for improving the performance of object instance recognition.

% Universal Correspondence Networks~\cite{choy2016universal} defines a framework for visual correspondences.
% % Apart from dense geometric correspondence,
% But it also focuses on keypoint \textit{semantic correspondences}.
% % for example, matching the eyes of one cat to those of another cat of a different breed.
% Dense Object Nets~\cite{florence2018dense} predict dense correspondences between instances in different viewpoints indirectly by learning dense pixel descriptors. However, they do this via self-supervision using a calibrated robot and dense 3D reconstruction.
% FlowNet~\cite{dosovitskiy2015flownet} and FlowNet2~\cite{ilg2017flownet} are popular, state-of-the-art deep learning systems to compute \textit{optical flow} between two consecutive frames of a video.
% % We build upon the basic architectures of FlowNet to compute dense pixel correspondence between the template image and scene images with some necessary modification (more details in Section \ref{sec:approach-flow}).
% We chose to customize and build upon this framework because \citet{dosovitskiy2015flownet} show that it is possible to train good optical flow models relying only on synthetic data just from template images.
% % However, we note that our method can incorporate any dense pixel correspondence framwork.

% \textbf{Neural Network Verifiation:}
%, in the sense that not verifying a detection is similar to rejecting it

\textbf{Classification with reject option.}
We use a set of verification tests that try to verify a detection and an object class that is proposed by an initial object detector.  
This notion of verification is related to the literature on classification with rejection.  Some of these previous approaches assume a cost function for rejection is provided and find the optimum rejection rule \cite{chow1970optimum,bartlett2008classification,cortes2016boosting} or learn classification and rejection functions simultaneously~\cite{geifman2019selectivenet,cortes2016learning}. Some works \cite{shafer2008tutorial,hechtlinger2018cautious} treat the problem as conformal prediction, where the classification system can predict any subset of classes, including the null set which stands for rejection.
In contrast, in our work, we verify whether the object class output by the detector is correct, and we re-score the confidence of detections based on the verification tests.

\section{Theoretical Framework}
\label{sec:theoretical-framework}

\theoremstyle{plain}
\newtheorem{assumption}{Assumption}

\newtheorem{theorem}{Theorem}

\newcommand{\rfil}{\mathcal{F}}

In this section, we lay out a formal framework for high-precision instance detection. We specify a set of assumptions and prove that under these assumptions, the verification-based detector will have no false positives.
We then discuss a modification of this method that we implement in practice which improves overall detection accuracy.

\textbf{Problem Statement}. We assume there are $O$ categories of objects that we are trying to recognize.  We also assume access to a dataset which consists of a variety of images per object, recorded from a set of different viewpoints and lighting conditions. We refer to these images as ``template images" (see Figure~\ref{fig:pipeline}). At test time, we are given ``scene images" -- these are real world images that contain (multiple) objects that need to be detected (see Figure~\ref{fig:pipeline}). We also assume that we have an object instance detector trained to detect objects of interest; this detector will return detections of the target objects, along with a proposed object class for each detection. However, many of these proposed object classes will be incorrect, i.e. false positives.  The goal of our framework is to filter these out.

%an index set $\mathcal{I}$ of objects.
% $\mathcal{V}_i$ is the set of viewpoints of $i$th object.

%, of a rigid-body transformation

%, of a rigid-body transformation
%An 2D image of a rigid object that is viewed from a different viewpoint corresponds to such a transformation.

%is the confidence score of a base instance detector system when matching $O_{i,m}$ and $D$.

%, and we denote $I_{i, m}$ as a rendering of the $m$th viewpoint of object $O_i$. 

\textbf{Notation}. We denote by $O_i$ the $i$th object from the $O$ categories of objects.   For each object $O_i$, we assume access to a dataset of $M_i$ template images recorded from a variety of viewpoints and lighting conditions; we denote $I_{i, m}$ as the $m$th template image for object $O_i$.
We denote by $T : (u_1,v_1) \to (u_2, v_2)$ a 2D mapping from pixels of one image to the pixels of another image.  We overload notation such that $T(I)$ is an application of the 2D mapping $T$ to all pixels in the image $I$ to produce a new image $T(I)$.  Let $r(T)$ denote the ``rigidity" of such a transformation, measured by the fraction of inlier pixel matches under the best approximating rigid transformation (see Appendix~\ref{supp:theoretical-framework}).  We denote by $T^R$ the set of such 2D mappings which are perfectly rigid, such that $r(T) = 1$. 
Let $D$ denote a detection in a scene image (e.g. a crop of a scene image). %Note that, in the theoretical framework, we are implicitly assuming that all detections contain an image of some object in our dataset.

%; let $gt(D)$ be the corresponding ground-truth object class of detection $D$

We assume access to a similarity classifier which returns a score $c(I_1, I_2)$ indicating the confidence that images $I_1$ and $I_2$ each contain the same object class.
We also assume access to a distance metric $d(I_1, I_2)$ that measures the the image-distance between images $I_1$ and $I_2$; this could be any distance metric that satisfies certain properties specified in Appendix~\ref{supp:theoretical-framework} such as  pixelwise $L2$-distance or normalized cross-correlation.
We make the below assumptions under which we will prove that our algorithm will lead to no false positives. We will need to relax these assumptions for a practical implementation; nonetheless, this framework provides the theoretical basis for our approach.

% $F(O_{i, m}, D)$ is a predicted mapping between target image $O_{i,m}$ and detection $D$. The result is a transformation $T$.
% Our flow-based dense pixel correspondence method plays the role of $F$.

% We assume access to a base object classifier, which returns a score $c(O_{i,m}, D)$ that indicates the confidence that detection $D$ is an image of the same object class as the object in image $I_{i,m}$ (class $O_i$).  

%detection $D$ is an image of the same object class as the object in image $I_{i,m}$ (class $O_i$).

% $\mathcal{F}(O_i, D) = m : \exists T^R \text{s. t. } d(T^R(O_{i,m}), D_j) \leq \gamma$; it is the set of all template viewpoints of $O_i$ can be rigidly transformed to $D_j$ within a perceptual tolerance of $\gamma$.

% \begin{assumption}
%     (\textbf{Generative Process of detections})\\
%     Every detection $D_i$ is generated by sampling a viewpoint of object $O_i$, rendering an image at this viewpoint, and applying a lighting variation within a distance $\gamma$ of the rendered image (a more formal definition is given in Appendix \ref{supp:theoretical-framework}).
% \end{assumption}

\begin{assumption}
    \textbf{(Dense Dataset)}\\
    \label{as:denseData}
    %There are enough number of template viewpoint images of $O_i$ available such that for a true detection
    For each detection
    $D$, which is of class $O_i$, $\exists m \in M_i, T \in T^R$ such that $d(T(I_{i, m}), D) \leq \gamma$.
    \end{assumption}
    
\begin{assumption}
    \textbf{(Similarity Classifier Smoothness)}\\
    We have a similarity classifier $c$ with a corresponding constant $\delta$ that satisfies the following property: for any two images $I_1$ and $I_2$, and for any detection $D$, if $\exists T \in T^R$ such that $d(T(I_1), I_2) \leq 2\gamma$, then $|c(I_1, D) - c(I_2, D)| < \delta$.
\end{assumption}
    
%$i, j, m, n$ and $D$ generated according to assumption 1$

% \begin{assumption}
%     \textbf{(Classifier Smoothness)}\\
%     We have an object classifier $c$ that satisfies the following property: for any two images $I_{i,m}$ and $I_{j,n}$ of object classes $O_i$ and $O_j$ respectively, and for any detection $D_k$  of object class $O_k$ satisfying assumption 1, if $\exists T \in T^R$ such that $d(T(I_{i, m}), I_{i, n}) \leq 2\gamma$, then $|c(I_{i, m}, D_k) - c(I_{i, n}, D_k)| < \delta$.
% \end{assumption}

%The first assumption states that detections are generated from objects in our dataset with viewpoint variation and a bounded amount of lighting variation; this assumption implicitly ignores occlusions or background variation, although our practical implementation handles such cases. 

The first assumption states that the dataset is dense enough such that every detection can be constructed as a rigid transformation of some template image of that object in the dataset, with a bounded lighting change applied.  Note that this assumption implies that all detections contain an image of some object in our dataset. The second assumption states that the similarity classifier is a smooth function: if two images $I_1$ and $I_2$ are sufficiently similar such that $I_2$ can be created by a rigid transformation and a small lighting change applied to $I_1$, then the similarity classifier $c$ will output a similar score when comparing $I_1$ and $D$ as when comparing $I_2$ and $D$.  %Note that we make no other assumptions on the output of the similarity classifier $c$, and hence this similarity classifier is not, by itself, capable of creating the strong results that we will provide below for our overall system; it must be combined with our approach for verifying dense correspondences.

%Note that an image $I_i$ which contains an object of a different class from the ground-truth class for detection $D_j$ might still output an arbitrarily high similarity score $c(I_i, D_j)$.

%a false positive dethere might be two object classes $i$ and $j$ that output a sufficiently high similarity score 

%can be described by a rigid 

%two inputs (viewpoints of different objects) are similar, then they will produce similar output (confidence scores for detections) at test time.
% (analogous to \finlier) 
%$r(\hat{T})$
%confidence of the one-shot instance detector for the current template image versus all other images. 

Our object detection pipeline proceeds as follows: we assume that an initial detector finds detections $D$ in an image and proposes an object class $O_i$ for each detection.  We then pass the detection and its corresponding proposed class to our verification system \textsc{TheoreticalFlowVerify}($i, D$) which returns True if it can verify that  $D$ contains an image of class $O_i$ and False otherwise.  Note that $\textsc{TheoreticalFlowVerify}$ may return False either because $O_i$ is the wrong object class or because the algorithm is simply unable to verify the class of the object.  However, as we will show, the key to our approach is that $\textsc{TheoreticalFlowVerify}$ will only return True when $O_i$ is the true object class; hence when the system returns True, we can rely on the detection being accurate.

We present $\textsc{TheoreticalFlowVerify}$ in detail in Appendix \ref{supp:theoretical-framework}; this method is a theoretical version of our practical algorithm \flowverify\ described in Section~\ref{subsec:verification-tests}. The method iterates over all template images $I_{i,m}$ of the proposed object class $O_i$; it then estimates a set of dense pixel-wise correspondences $\hat{T}$ between the detection $D$ and the template image $I_{i,m}$.
It then determines whether it can validate that $D$ is of class $O_i$ based on image $I_{i,m}$. These verification tests are: 
\begin{enumerate}
    \item Similar Object Comparison: Tests whether an image $I_{j,n}$ of another object class $O_j$ looks similar to $D$ according to similarity function $c$; formally:
    $\forall j \neq i, \forall n \in M_j$, if $c(I_{i,m}, D) < c(I_{j,n}, D) + \delta$, return False
    % \item Similar Object Comparison: The difference in output between the similarity classifier $c$ for a template image $I_{i,m}$ of the proposed object class $c(I_{i,m}, D)$ and images $I_{j,n}$ from all other classes $c(I_{j,n}, D)$
    \item Color Comparison: Tests whether the image distance between $I_{i,m}$ transformed using $\hat{T}$ and the detection $D$ is sufficiently small: if $d(\hat{T}(I_{i,m}), D) > \gamma$, return False
    \item Flow Rigidity: Tests whether the the correspondences $\hat{T}$ could have been derived from a rigid object transformation: if $r(\hat{T}) < 1$, return False
\end{enumerate}

% between a template image and a detection
%$I_{i,m}$

% (1) an estimate of the rigidity  of a predicted set of correspondences  $\hat{T}$ between a template image and a detection, (2) , and . The algorithm either accepts the detection and returns its class-id or rejects it by returning \texttt{Unknown}.
% It can sometimes fail to classify detections and reject them.
%We call this algorithm \textsc{TheoreticalFlowVerify}.

% For this algorithm, we can prove the following theorem:
\begin{theorem}
    \textbf{(No False Positive Theorem)} [Proof: see Appendix~\ref{supp:theoretical-framework}]\\
    Under assumptions 1 and 2, \textsc{TheoreticalFlowVerify} does not produce any false positives. That is, the following statement always holds: $\textsc{TheoreticalFlowVerify}(i, D)$ returns False whenever the ground-truth class for detection $D$ is different from $O_i$.  
\end{theorem}

%    
    %True only if the ground-truth class for detection $D$ is class $i$.  
%    $\textsc{TheoreticalFlowVerify}(D_i) \in \{i, \texttt{Unknown}\}$.

% \begin{proof}
%     See Appendix~\ref{supp:theoretical-framework}.
% \end{proof}

%also discuss how our framework is more general and incorporates other comparision functions and families of transformation. 
%    That is, \textsc{TheoreticalFlowVerify} always either returns the true object class or returns \texttt{Unknown} and so 

Note that $\textsc{TheoreticalFlowVerify}(i, D)$ returns False whenever it cannot verify that the ground-truth class of $D$ is $O_i$.  This can sometimes occur 
even if $O_i$ represents the ground-truth class of $D$, if the estimated correspondences $\hat{T}$ are not accurate.  It will also return false whenever the ground-truth class of $D$ is not $O_i$.  Thus \textsc{TheoreticalFlowVerify} has 100\% precision; every time it returns True, the proposed object class is the same as the ground-truth.  On the other hand, the recall of the algorithm is not guaranteed; it may return False for an arbitrary proportion of examples, even if the proposed object class is correct.  Note that the No False Positive Theorem does not make any assumptions on the quality of the set of predicted correspondences $\hat{T}$; if the correspondences are not accurate, then \textsc{TheoreticalFlowVerify} will also reject the detection.
% We will now describe an approximation of this framework that leads to a practical implementation of this algorithm. 

% Note that we do not make any \textit{explicit} assumptions about the image acquisition process  for scene or template images. However, they may be implied by Assumptions 1 \& 2 as stated above. For example, these assumptions disallow large lighting changes or very low resolution imaging that could generate false positives that fool our verification tests. That is, it is not possible for an object detected in the scene to look more similar to another object, get detected as such, and pass all verification tests.

One might wonder how it is possible to guarantee no false positives, when no \textit{explicit} assumptions are made about the image acquisition process. However, this concern is resolved by assumption 1, which gives constraints on the detection images that imply constraints on the image acquisition process.  Assumption 1 states that, for each detection, there exists some template image of the true object that is sufficiently similar in appearance to the detection image. This assumption thus disallows, for example, large changes in resolution between the detection image and its corresponding template image. Without assumption 1, an object recorded at a low resolution might look more similar to a different object's template and create a false positive.

%; it can determine that the correspondences are not accurate by measuring the image distance after applying $\hat{T}$ to the template image and comparing to the detection, while making sure that $\hat{T}$ is rigid and the detector is sufficiently confident

%In Appendix~\ref{supp:theoretical-framework}, we describe various generalizations of this framework, and in the next section, we will describe an approximation of this framework that leads to a practical implementation.  

%For our experiments, we implement \flowmatchnet, our FlowNet based dense pixel correspondence as our predicted transformation $\hat{T}$; note that the perfect theoretical precision of \flowverify\ does not rely on the accuracy of the correspondence. We implement $\finlier$ for $r(T)$. We found that L2 and other similar perceptual similarities do not work well in practice. And instead of relying on the base detector to produce large differences in confidence while predicting the correct class of the detection, we propose other verification tests ($\fp$ and $\fr$ --  see Section \ref{subsec:verification-tests}) that work well in practice.
%\sid{maybe we should better justify the modification in tests we're doing?}

\vspace{-5pt}

\section{Approach}

We will now describe an approximation of the theoretical framework that leads to a practical implementation of our method. The pipeline for high-precision detection consists of three stages:

\begin{enumerate}
   \item \emph{Base Instance Detector}: We first run an instance detector trained on the objects of interest.
   This stage provides candidate bounding box detections for target objects, as well as a proposed object class $O_i$ for each box. 
   The focus of our work is verifying these detections, as described below.
   %in 2 and 3, to increase precision and overall accuracy of the detector. Our method could be used with any instance detector.
   
   \item \emph{Dense Pixel-wise Correspondence} (\flowmatchnet): We next predict dense pixel-wise correspondences between template images and each proposed detection, cropped from the scene. See Appendix \ref{sec:approach-flow} for more details on the network architecture of \flowmatchnet.
   
   \item \emph{Verification tests} (\flowverify): Given the proposed detection and template images and estimated pixel-wise correspondences between the two, we conduct a set of \textit{verification tests} to ascertain whether the proposed class of the detection is correct.
\end{enumerate}

%   If the proposed detection passes the tests, it should be a true positive with very high probability.
%   We assume that false negatives are somewhat tolerable.

\subsection{Verification Tests}
\label{subsec:verification-tests}

We design verification tests that are modifications to our theoretical framework (Section~\ref{sec:theoretical-framework} and Appendix~\ref{supp:theoretical-framework}).
These tests, which we call `\flowverify' tests (since they make use of predicted flow correspondences), are intended to be stringent; any detection that does not pass these tests is deemed `rejected' and will receive a lower detection score than the detections that pass the tests.
%The hope is that most false positive detections will be rejected by these tests.
% Some true positives may be rejected  as well; we consider this to be an acceptable tradeoff to create a high precision detector.
We can be highly confident that the detections that pass our verification tests are likely to be true detections, boosting the performance of our detector in the high-precision regime, as our results demonstrate.
%We design five verification tests using pixel-wise correspondences between the template image and proposed detection:
The first three verification tests are derived from our theoretical framework of Section~\ref{sec:theoretical-framework}:

\textbf{1. \nms}: This test corresponds to the similar object comparison test in our theoretical framework. For each detection, there should be only one target object being matched to it with high confidence. The similar object comparison test is the following: for each detection $D$ we compute a confidence score $c(D)$ using the initial detector (such as TDID~\cite{tdid}).  We then search whether there is another detection $D'$ of another object which has an IoU of at least $\eta_{iou}$ with $D$. If no such detection is found, $\nms = 1$. Otherwise, the similarity score for $D$ is the minimum of the confidence difference across all other detections $D'$: $\nms\ = \min_{D'} \max(0, c(D) - c(D'))$.  We define the \textit{similar object} test using a boolean variable given by $T_{\nms} = (\nms > \eta_{diff})$.

%check whether $c(D') /c(D)\geq \eta_{ratio}$. If such $D'$ exists, \nms\ = 0; otherwise \nms\ = 1. We define the \textit{similar object} test using a boolean variable given by $T_{\nms} = (\nms =1)$.

%the \textit{similar object} test as

% $I_{i,m}$ transformed using $\hat{T}$ and the detection $D$ is sufficiently small: if $d(\hat{T}(I_{i,m}), D) > \gamma$
%If the cropped detection $D$ contains the object, our assumption \ref{as:denseData} holds, and \flowmatchnet computes accurate 

% \textbf{2. \fcolor}: This test corresponds to the color comparison test in our theoretical framework. We estimate the pixel-wise correspondences $\hat{T}$ between a template image $I_i$ of the proposed object class $O_i$ and the detection $D$.  We then check the image similarity between $D$ and the template image transformed using the predicted correspondences $\hat{T}(I)$.  We use normalized cross correlation (``ncc") to measure this similarity. We define $\fcolor = \text{ncc}(\hat{T}(I_{i}), D)$ and we define
% the \textit{flow color} test using a boolean variable given by 
% $T_{color} = (\fcolor > \alpha_{color})$.

\textbf{2. \fcolor}: This test corresponds to the color comparison test in our theoretical framework. We estimate the pixel-wise correspondences $\hat{T}$ between a template image $I_i$ of the proposed object class $O_i$ and the detection $D$.  We then check the image similarity between $D$ and the template image transformed using the predicted correspondences $\hat{T}(I_i)$.  We use normalized cross correlation (``ncc") to measure this similarity, which lies in $[-1, 1]$. We define $\fcolor = \frac{1}{2}(\text{ncc}(\hat{T}(I_{i}), D)+1) \in [0, 1]$ and we define
the \textit{flow color} test using a boolean variable given by 
$T_{color} = (\fcolor > \alpha_{color})$.

\textbf{3. \finlier}: This test corresponds to the rigidity test in our theoretical framework.
% where we check the computed flow corresponds to a rigid transformation of the object between the dataset and the scene image.
%and \flowmatchnet\ computes perfectly accurate pixel-wise correspondences between template image and the bounding box prediction, we would 
%This can be described by the fundamental matrix.
Assuming objects are rigid, if the detection box contains the target object, then we would expect an ideal mapping between the detection and a corresponding template image from a similar viewpoint to describe a rigid body transformation. 
We use RANSAC~\cite{fischler1981random} to find the best-fit fundamental matrix (using the 8-point algorithm~\cite{longuet1981computer}) that maximizes the number of corresponding inlier pairs.
The proportion of inliers under the best-fit mapping is a measure of how rigid the flow is. We define flow rigidity as
% For the predicted flow, we use RANSAC~\cite{fischler1981random} to select a set of find a homography that maximizes the number of pixel correspondences which satisfy the epipolar constraint - in other words, we find the homography that maximizes the number of inliers (where an inlier is a corresponding pair of pixels that satisfies the epipolar constraint). We define the \textit{flow homography inlier ratio} as -
\[\finlier = \frac{\text{\#inliers }}{\text{\#total correspondences}}  \]
% If detection box and flow are accurate, flow rigidity inlier ratio would equal 1.
and we define a \textit{flow rigidity} test using a boolean variable $T_{rig} = (\finlier >\alpha_{rig})$. Besides the verification tests motivated by our theoretical framework, we design two additional tests to evaluate the extent of each bounding box prediction: 

\textbf{4. \fp}: If we expect the detection box to contain the entire object (and not be too small), we would expect all pixels in the template image to be mapped to pixels \textit{inside} the bounding box. We define \textit{flow precision}  as
\[\fp\ = \frac{\text{\#target correspondences mapped to inside bbox}}{\text{\#total correspondences}} \]
and the \textit{flow precision} test is defined using a boolean variable $T_{prec} = \left(\fp  > \alpha_{prec} \right)$.

\textbf{5. \fr}: Complementary to precision, if we expect the detection box to contain only the detected object (and not be too large), we would expect the flow mapping to cover all pixels of the detection box. To measure this, we can look at the bounding box that tightly fits the extent of the mapped pixels in the scene. We define \textit{flow recall} as the IoU between the box suggested by flow to the  bounding box output by the detector, and compute a \textit{flow recall} score as 
\[\fr = \text{IoU}(\text{detector bbox}, \text{bbox suggested by flow}).\]
The \textit{flow recall} test is defined using a boolean variable $T_{rec} = \left( \fr\ > \alpha_{rec}\right)$.
% Note that \nms, \fcolor, \finlier, \fp, and \fr all lie in the range $[0, 1]$. Then we have five tests $T_{\nms} = (\nms =1)$, $T_{color} = (\fcolor > \alpha_{color})$, $T_{rig} = (\finlier > \alpha_{rig}), T_{prec} = (\fp > \alpha_{prec})$ and $T_{rec} = (\fr > \alpha_{rec})$, 
Our overall verification test is 
\[\flowverify\ = T_{\nms}\ \wedge T_{color}\ \wedge T_{rig}\ \wedge T_{prec}\ \wedge\ T_{rec},\]
with given threshold values and parameters $\eta_{diff}, \alpha_{color}, \alpha_{rig}, \alpha_{prec}, \alpha_{rec}  \in [0, 1]$, tuned with a validation set as described in Section~\ref{sec:experiments}. % and \nms\ parameters $\eta_{iou}, \eta_{ratio} \in [0, 1]$.
For a given object of interest, we could have multiple template images from different viewpoints and lighting conditions. In this case, we say that the set of template images for a given class passes \flowverify\ if any one of the template images for that class passes these tests.
%The motivation is that any template image that is of the correct viewpoint as the object in the scene, is expected to produce a match.
% Whereas other viewpoints, say the rear- or side- views, are not expected to match.
%In order to find the best values of $\eta_{diff}, \alpha_{color}, \alpha_{rig}, \alpha_{prec}, \alpha_{rec}$, we perform a grid-search over a validation set to optimize precision of the top scoring detections.
%We show that the performance generalizes to novel, unseen test objects in Section \ref{sec:experiments}.

While \flowverify\ tests are designed to improve performance in high precision regime, they might worsen performance in the low-precision regime as they could reject true positives. As there are cases where performance in low-precision regime is also important, instead of completely rejecting detections not passing  \flowverify\ tests, we rerank all detections from our base detector based on 1) whether a detection passes \flowverify\ tests and 2) the score predicted by base detector. All detections that pass \flowverify\ are ranked higher than all detections that don't. The reranking procedure can be viewed as reducing the confidence of the detections that do not pass the \flowverify\ tests. As we will see, this improves performance in the high-precision regime while at the same time maintaining performance in the low-precision regime.

\section{Experiments}
\label{sec:experiments}

\textbf{Datasets}.
We evaluate our framework on two tests sets -- the GMU Kitchens test split~\cite{gmuk} and W-RGBD scenes v1 Dataset \cite{lai2014unsupervised}. Both datasets contain images of objects placed in indoor scenes such as kitchen surfaces, table tops, and living rooms.  For \flowmatchnet\ and \flowverify, at test-time we use 15 equally spaced viewpoints  per object taken from~\cite{bigbird} or \cite{5980382} to form our ``template images".
We also vary the number of template images to explore the speed-accuracy tradeoff of our system in section~\ref{sec:main-result}.
% These would come from a dataset of objects.
For GMU, we evaluate on the 11 BigBIRD objects; for W-RGBD, we evaluate on 9 textured objects. 
% We conduct experiments for \flowmatchnet\ and \flowverify\ on three instance detection datasets -- Yale-CMU-Berkeley (YCB) Dataset~\cite{calli2017yale}, Active Vision Dataset (AVD) \cite{avd}, and GMU Kitchens \cite{gmuk}. Each dataset contains images of objects placed in indoor scenes such as kitchen surfaces, table tops, living rooms etc. 

% The YCB-Video train dataset is used as a validation set for finding hyperparameters of \flowverify. The AVD dataset is used as a training set for our base instance detector, and the GMU Kitchens test split is used for evaluation. As AVD and GMU Kitchens share objects from the BigBIRD~\cite{bigbird}, the shared objects are removed from AVD during training.

\textbf{Base Instance Detector} In this work, we focus on improving the precision of an existing instance detector. One could use any instance detector with our approach; we use  \textit{target driven instance detection} (TDID)~\citep{tdid}, as this method produces  state-of-the-art results for instance detection, including one-shot instance detection that we evaluate on in our work. In the one-shot scenario, the training and validation objects are separate from the objects evaluated on at test time; the detector must generalize to novel objects.  TDID showed state-of-the-art performance for this task~\citep{tdid}.  Similarly, we train other components of our system, such as \flowmatchnet , in a one-shot manner, such that our entire system can be used to detect novel objects.

%As this base detector is \textit{one-shot}, it can detect instances of novel objects not seen during test time. Since other components of our pipeline are also one-shot, we expect our entire method to generalize to novel objects, and it can be used as-is to detect novel objects in different environments while obtaining high precision.

The base detector TDID is trained on the Active Vision Dataset (AVD) \cite{avd}, W-RGBD scenes Dataset \cite{lai2014unsupervised}, and synthetic images from Cut-Paste-Learn \cite{dwibedi2017cut}. GMU objects common to the AVD dataset are removed from training so that we can evaluate this detector in a one-shot manner.
In order to evaluate on W-RGBD Scenes dataset in a one-shot manner as well, for this evaluation we use the TDID model released by \citep{tdid} that is trained only on AVD. 

\textbf{\flowmatchnet}. Similar to previous optical flow models~\cite{dosovitskiy2015flownet,ilg2017flownet}, we train \flowmatchnet\ on synthetic data for which ground-truth correspondences can be computed. We train \flowmatchnet\ successively on synthetic datasets of increasing difficulty. First we train on  objects from MS-COCO~\cite{lin2014microsoft} with synthetic affine transformations; next we train on  BigBIRD~\cite{bigbird} images that are cropped out and blended into background via homographic transformations,  randomized lighting conditions, and image blurs. 
More details are in Appendix~\ref{supp:training-flowmatchnet}.

%
%(all detections with a score above a threshold of 0.999)

%(i.e. the number of true positives dominate the total number of detections above a certain threshold)
%As we hope the method to generalize across different environments, w

\textbf{\flowverify}. We use a validation set to tune the parameters for the tests in \flowverify.
% Since we wish to create a detection system that operate in the high precision regime
%We optimize thresholds for maximizing overall performance for each object.
%To do so, we We employ grid-search on the 
As our validation set, we use the YCB-Video training set, on which we tune the values for $\alpha_{rig}, \alpha_{color},  \alpha_{prec}, \alpha_{rec},\eta_{diff} \in [0, 1]$ by optimizing mAP. We set $\eta_{iou}=0.5$, which is the common threshold used for mAP evaluation. We find the optimal values to be $\alpha_{rig}=0.9, \alpha_{color}=0.5, \alpha_{prec}=0.9, \alpha_{rec}=0.3$ and $\eta_{diff} = 0.0$ (meaning that all other object classes should have strictly lesser confidence scores if detected).

%. We employ another grid search for $\eta_{diff}$, and find

%We search with an interval of 0.1 for the $\alpha$'s one at a time, and

%Since we evaluate with IoU threshold of $0.5$, we set $\eta_{iou} = 0.5$.

\begin{figure}
  \centering
   \newcommand{\scale}{0.41}
  \begin{subfigure}[t]{\scale\linewidth}
  \includegraphics[width=\textwidth]{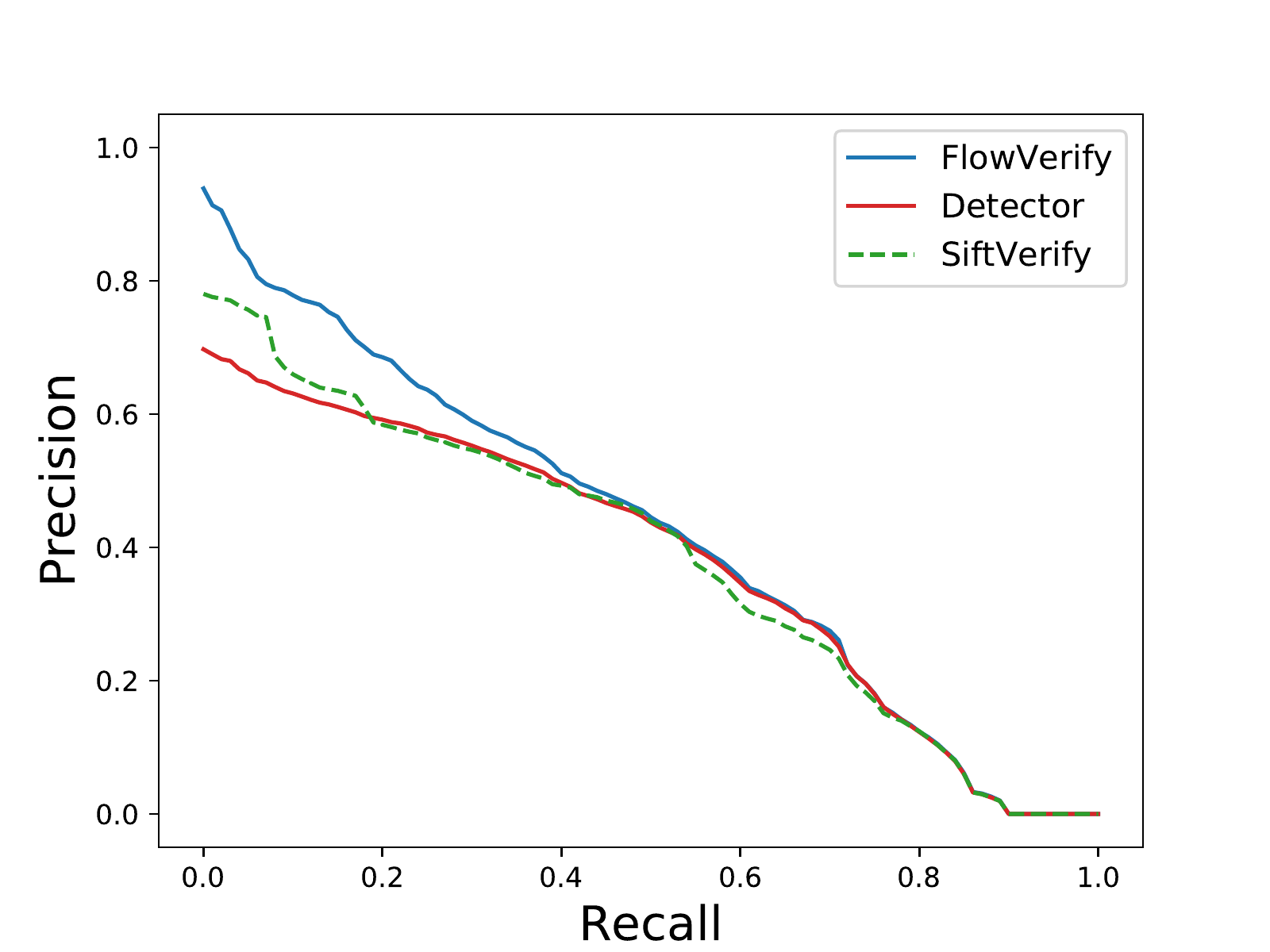}
  \caption{}
  \end{subfigure}
    \begin{subfigure}[t]{\scale\linewidth}
  \includegraphics[width=\textwidth]{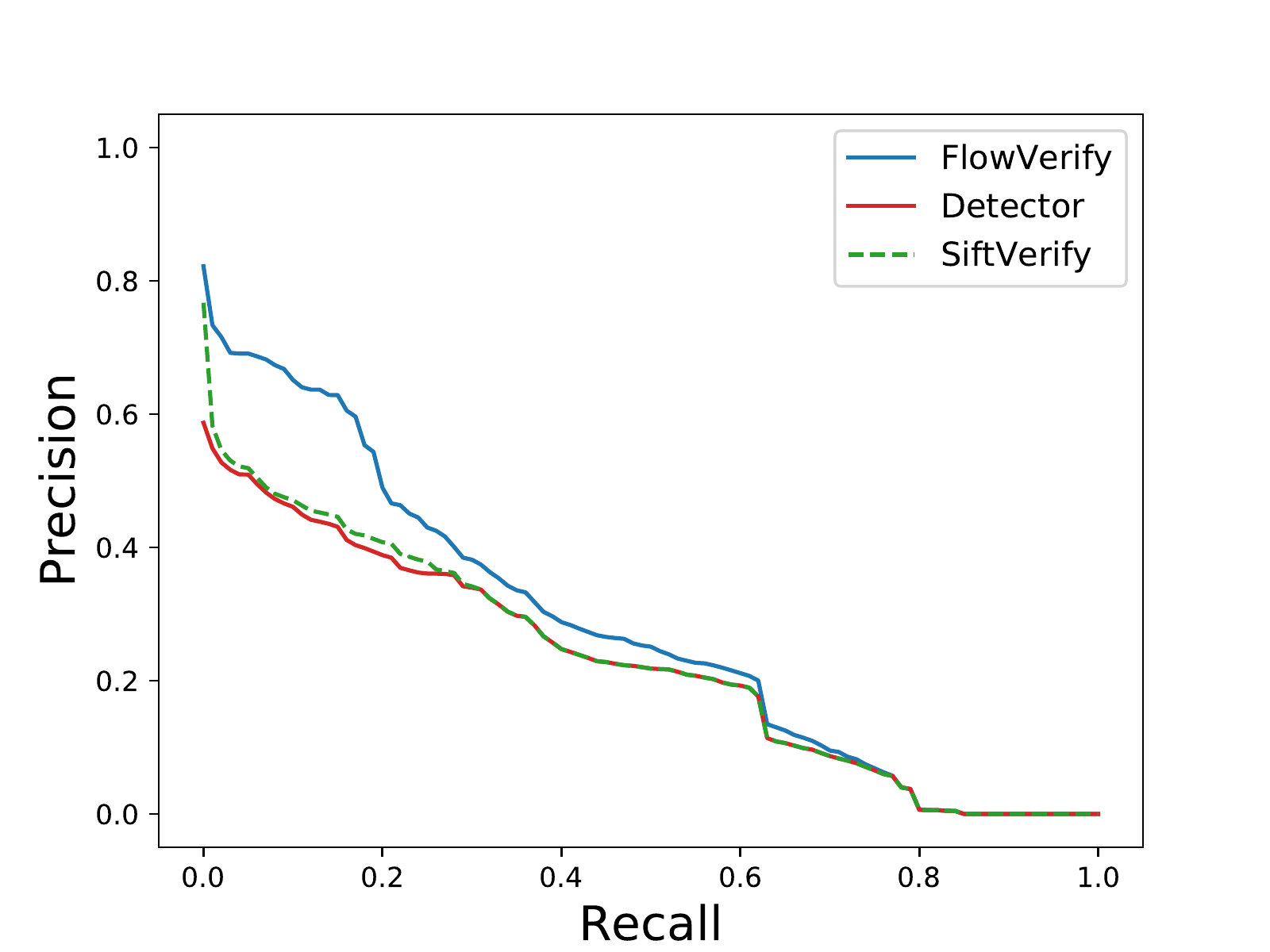}
  \caption{}
  \end{subfigure}
   \caption{Precision-Recall (PR) curves for the base instance detector, the detector improved by \flowverify, and the \siftverify\ baseline, evaluated on the GMU (a) and W-RGBD (b) datasets.}
  \label{fig:main_result}
  \vspace{-10pt}
\end{figure}

\subsection{Results}
\label{sec:main-result}
   Figure \ref{fig:main_result} shows precision-recall curves of the base instance detector (TDID) and the improved detector using \flowverify, on GMU (a) and W-RGBD (b) datasets. On both datasets, \flowverify\ significantly improves detection performance in the high-precision regime to the left end of the PR curve. For high scoring detections, it makes significantly fewer errors than TDID as it is able to filter out many false positive detections using the verification tests. Table \ref{wrap-tab:main-result} summarizes mAP and maximum precision of baseline detectors and \flowverify\ on GMU and W-RGBD. It is worth noting that although \flowverify\ is designed to reject false positives and improve performance in the high-precision regime, the PR curves show that, with the re-ranking procedure, the performance does not degrade in the low-precision regime (right side of the curve). In the low-precision regime, \flowverify\ nearly matches the base detector in performance. This shows that verification tests can make an instance detector more reliable and precise while simultaneously improving overall performance.
\begin{table}[htbp]
   \makebox[\textwidth][c]{
   \renewcommand{\arraystretch}{1.1}
   \begin{tabular}{c|cc|cc|c}
   \toprule 
   & \multicolumn{2}{c|}{GMU-Test} & \multicolumn{2}{c|}{RGBD} & \\\hline
   Method & mAP & max Precision  & mAP & max Precision & time(s) \\\hline
   %1.41, 1.14, 0.715, 0.357
   Base Detector              & 0.379    & 0.697 & 0.222    & 0.587 & \textbf{0.037}\\
   \siftverify                & 0.382    & 0.780 & 0.228    & 0.767 & 0.940\\
   \flowverify              & \textbf{0.422}    & \textbf{0.939} & \textbf{0.278} & \textbf{0.822} & 0.872\\
      \bottomrule
   \end{tabular}
   \renewcommand{\arraystretch}{1}
   }\\
      \caption{Overall performance (mAP) and max Precision results.}
   \label{wrap-tab:main-result}
   \vspace{-10pt}
\end{table}

We note that our method presents a tradeoff between precision/accuracy and timing performance. As more template images are used, the precision/accuracy of our system increases, but it will take longer to run. Hence, the user of our system can adjust the number of template images based on their specific needs. We report the running times of our method for varying number of template viewpoints in Table~\ref{wrap-tab:perf_vs_speed}.

\begin{table}[htbp]
   \makebox[\textwidth][c]{
   \renewcommand{\arraystretch}{1.1}
   \begin{tabular}{c|cc|cc|c}
   \toprule 
   & \multicolumn{2}{c|}{GMU-Test} & \multicolumn{2}{c|}{RGBD} & \\\hline
   Method & mAP & max Precision  & mAP & max Precision & time(s) \\\hline
   \flowverify (15 vp)             & \textbf{0.422}    & 0.939 & \textbf{0.278} & 0.822 & 0.872\\
   \flowverify-12vp          & 0.413    & 0.946 & \textbf{0.278} & 0.838 & 0.696\\
   \flowverify-9vp          & 0.411    & \textbf{0.962} & 0.275 & \textbf{0.847} & 0.540\\
   \flowverify-6vp          & 0.408    & 0.898 & 0.263 & 0.766 & 0.379\\
   \flowverify-3vp          & 0.405    & 0.932 & 0.252 & 0.783 & \textbf{0.201}\\
   \bottomrule
   \end{tabular}
   \renewcommand{\arraystretch}{1}
   }\\
   \caption{Tradeoff of performance vs speed as we vary the number of viewpoints.}
   \label{wrap-tab:perf_vs_speed}
   \vspace{-20pt}
\end{table}

%, including the base detector and \siftverify\ baseline

Table \ref{wrap-tab:new-ablation} in the Appendix shows an ablation analysis of our method; we remove verification tests in \flowverify\ one at a time and report performance on the test set. Dropping \finlier\ leads to the largest drop in performance in terms of both mAP and maximum precision.

\vspace{-8.5pt}

\subsection{\siftverify\ Baseline}
\label{subsec:sift-baseline}

We implemented and evaluated another verification method using SIFT-based~\cite{lowe1999object,lowe2004distinctive} keypoint correspondences as a baseline for our learning-based dense-correspondences computed by \flowmatchnet. This baseline is designed to test our hypothesis that verification tests should make use of both machine learning and non-machine learning approaches; we use machine learning for computing the correspondences (\flowmatchnet) but we use the non-learning based verification tests of \flowverify\ to verify detections. 

%Better aligned images typically yield a higher number of SIFT matches. 
%Hence we use a traditional keypoint matching approach to compute pixel correspondences.

In contrast, SIFT~\cite{lowe1999object,lowe2004distinctive} is a non-learning based approach for computing correspondences, and we combine it with a non-learning based approach for verification.
We implement verification tests on top of SIFT that are analogous to \flowverify, described in more detail in Appendix~\ref{sec:SiftVerify}. We call this baseline \siftverify. Figure~\ref{fig:main_result} and Table~\ref{wrap-tab:main-result} show the performance on the GMU and W-RGBD datasets. For both datasets, \siftverify\  slightly improves performance over the base detector, but performs consistently worse than \flowverify.  This demonstrates the value of using machine learning for computing correspondences, even if the final verification tests are not learned.
\vspace{-8.5pt}

%, but leads to either worse or almost similar mAP. This baseline 

%several correspondence lines criss-cross each other, indicating that the overall transformation is likely not 

\subsection{Qualitative Analysis}
\begin{figure}
  \centering
  \newcommand{\scale}{0.245}
  \begin{subfigure}[t]{\scale\linewidth}
  \includegraphics[width=\textwidth, trim={0 0 0 2cm},clip]{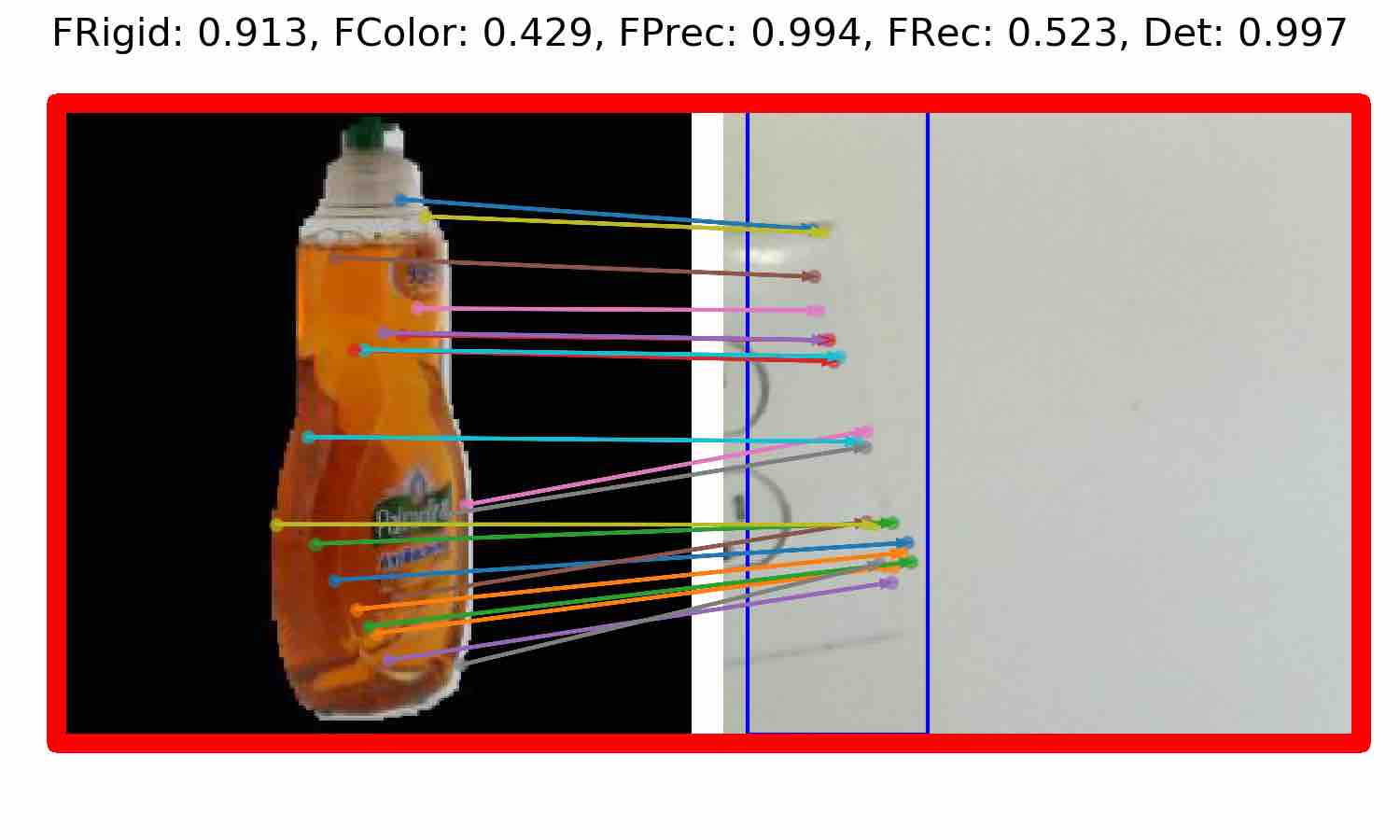}
  \caption{}
  \end{subfigure}
  \begin{subfigure}[t]{\scale\linewidth}
  \includegraphics[width=\textwidth, trim={0 0 0 2cm},clip]{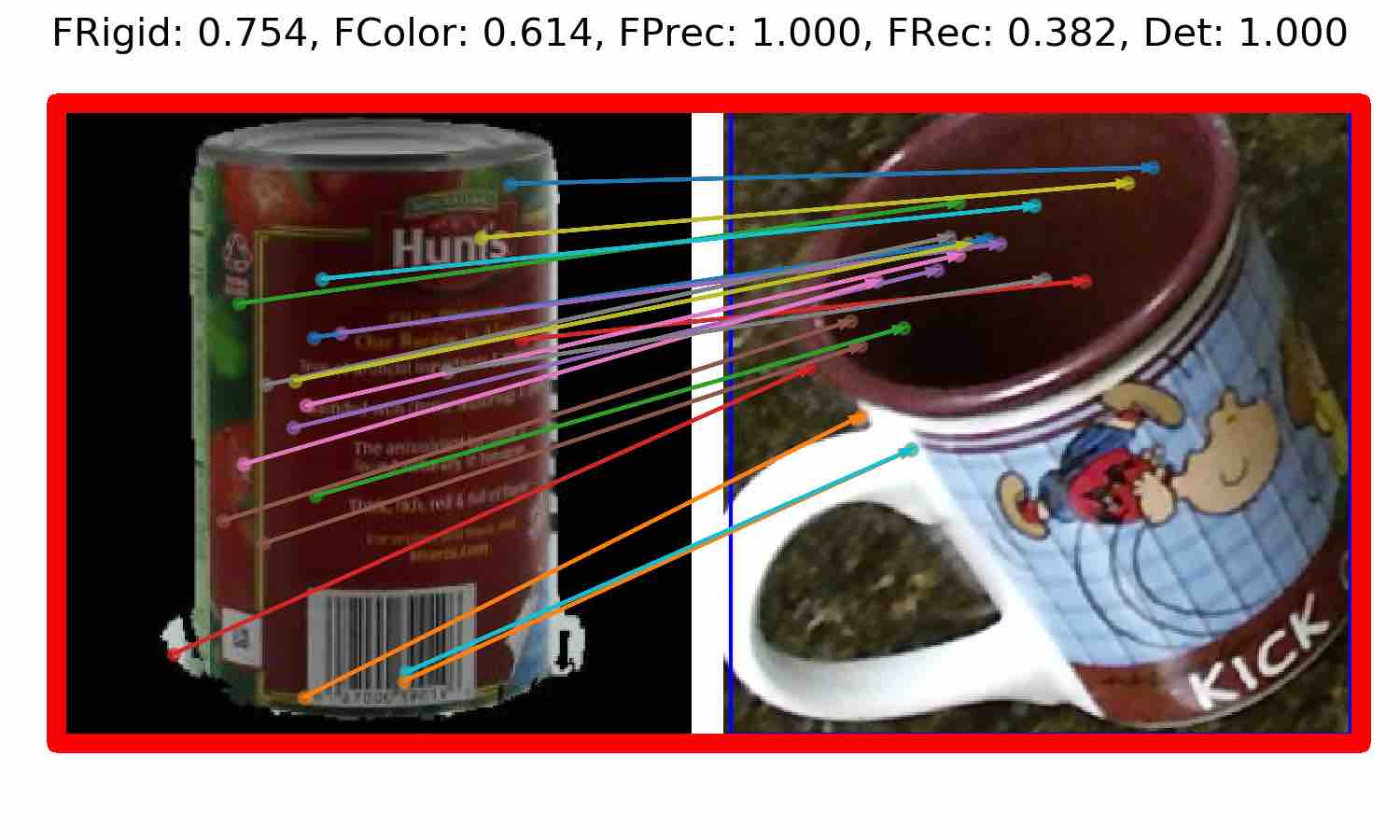}
  \caption{}
  \end{subfigure}
  \begin{subfigure}[t]{\scale\linewidth}
  \includegraphics[width=\textwidth, trim={0 0 0 2cm},clip]{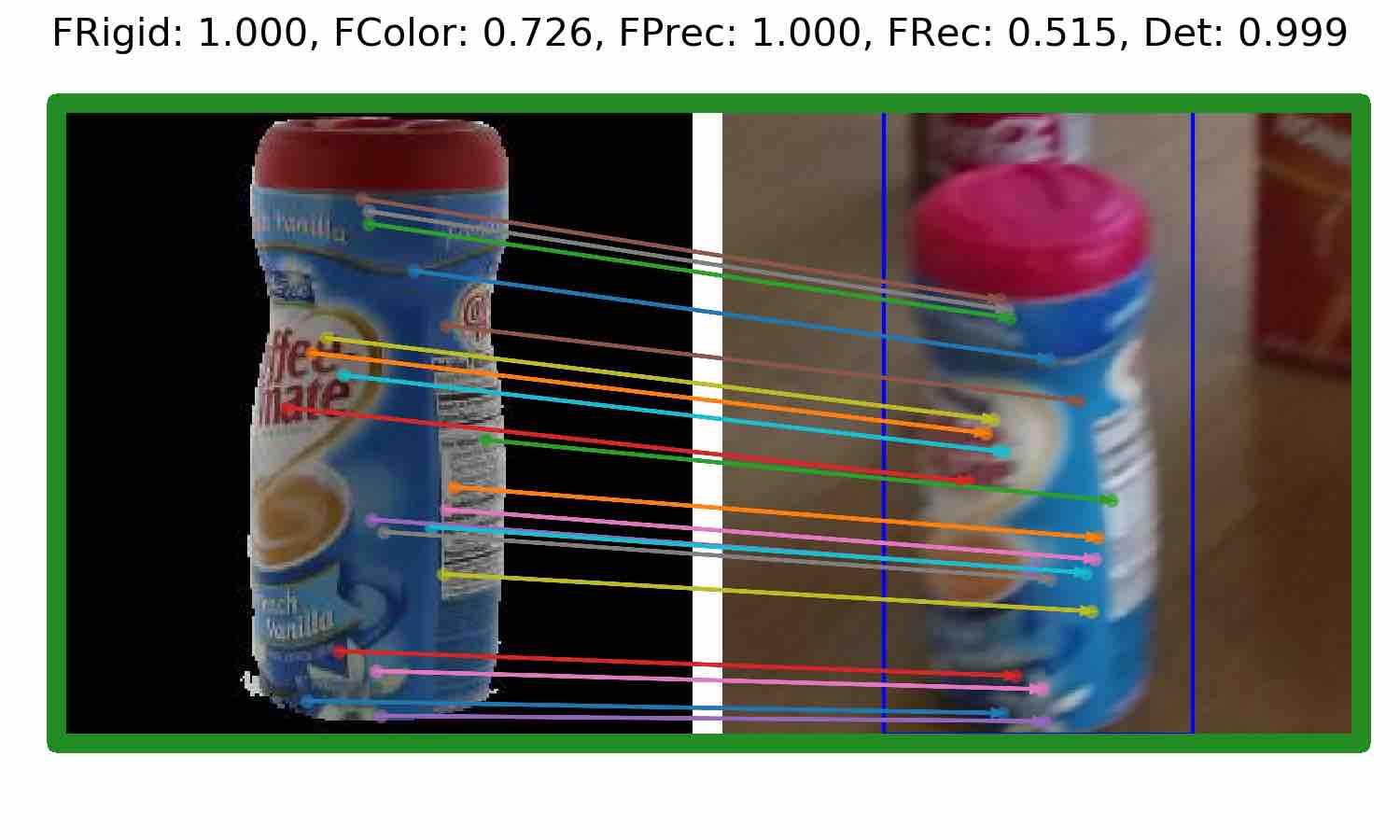}
  \caption{}
  \end{subfigure}
  \begin{subfigure}[t]{\scale\linewidth}
  \includegraphics[width=\textwidth, trim={0 0 0 2cm},clip]{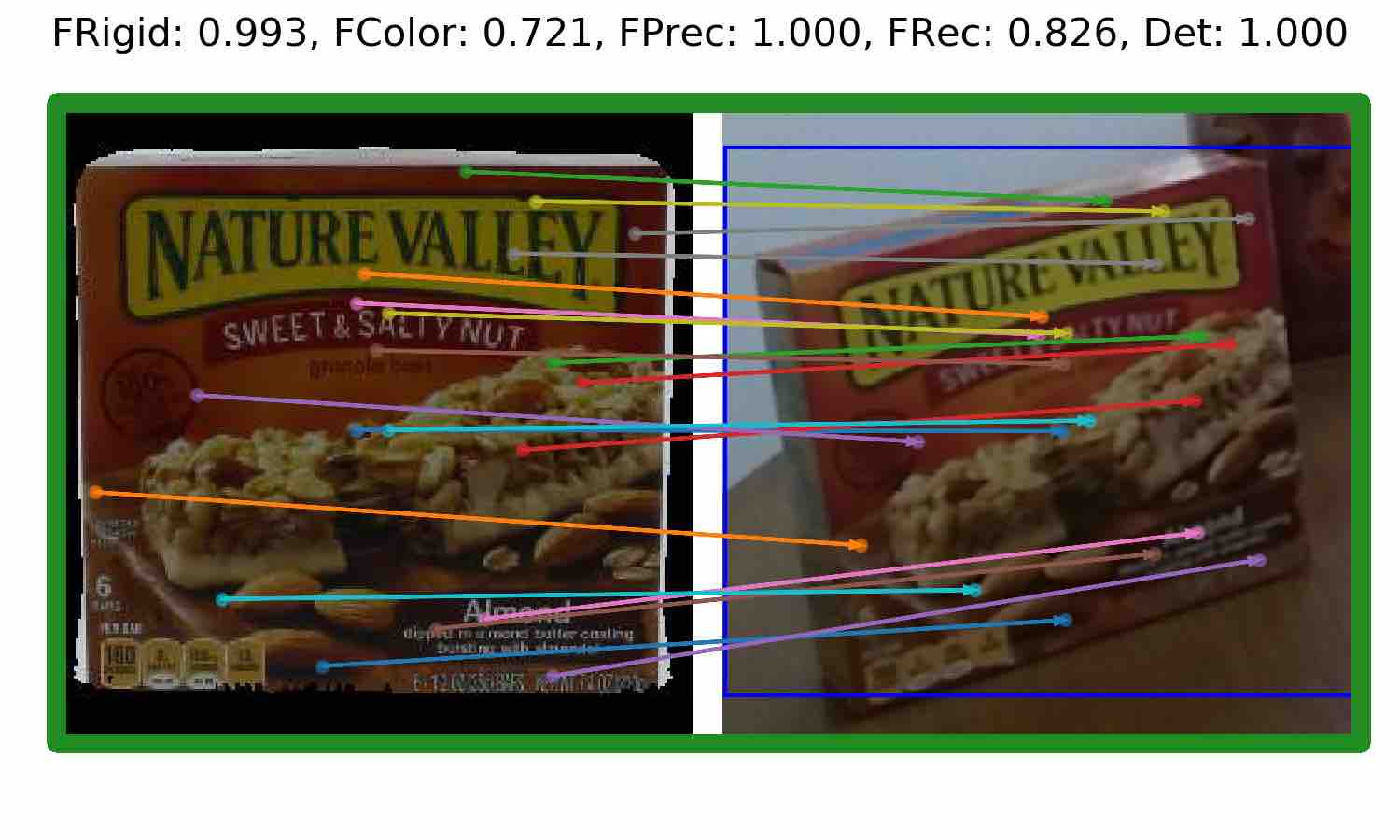}
  \caption{}
  \end{subfigure}
  \caption{(a, b): False positives from the base detector filtered by \flowverify; (c, d): True positives from the base detector accepted by \flowverify.}
 \label{fig:qualitative}
\vspace{-10pt}
\end{figure}

Figure~\ref{fig:qualitative} (a, b) shows examples of false positive detections output by the base detector (TDID) but filtered out using \flowverify. Example (a) is a detection with high \finlier , \fp , and \fr\ scores, but with a low \fcolor\ score. In such cases, the detection is of an object with very different color and texture from the target object, resulting in low color similarity.
Example (b) shows a detection with high \fcolor\ and \fp\ scores but with a low \finlier\ score. Here, the matched colors are fairly similar; however, the correspondences cannot be derived from a rigid body transformation, and hence the detection has a low \finlier\ score.
Examples (c) and (d) in Figure \ref{fig:qualitative} show true positive predictions output by the base detector which pass \flowverify. We can see that the flow quality is also quite good by the correct matching of features across the target and the bounding box area.

\vspace{-4pt}
\section{Conclusion}
\vspace{-2pt}
\label{sec:conclusion}
We have proposed a method to combine machine learning based detection and correspondence matching with non-learning based verification tests to increase the accuracy of an existing instance-detection system in the high-precision regime (without reducing overall detection performance).
The verification tests are based on dense-pixel correspondences computed between the detection and template images; we reduce the confidence of any detection that does not pass these tests, thereby rejecting many false positives.  Our system is grounded in a novel theoretical framework that we prove leads to no false positives, under certain assumptions.
Furthermore, we use our method in a one-shot fashion, applying our approach to a novel set of objects at test time without finetuning.
We hope that our system will be useful for robotic systems that need reliable performance for high confidence detections.

%Our system would especially be useful in helping these robots reject potentially unreliable detection, and then respond in a more reliable way.

% such as changing the viewpoint or asking a human for help.
%, that compare detections against template object images
%, thereby these tests are used to reduce the confidence of false positives
%We show that such a system can be perfectly precise in a theoretical setting.
%We show via experiments how the precision and accuracy of a state-of-the-art instance detection system can be improved.
%This approach generalizes to novel objects and requires no

%===============================================================================

% The maximum paper length is 8 pages excluding references and acknowledgements, and 10 pages including references and acknowledgements

% The acknowledgments are automatically included only in the final version of the paper.
\acknowledgments{This material is based upon work supported by the United States Air Force and DARPA under Contract No. FA8750-18-C-0092, the NSF under Grant No. IIS-1849154, as well as a Google Faculty Research Award.}

\clearpage

%NSF S\&AS grant 

%===============================================================================

\bibliography{example}  % .bib

\providecommand{\noopsort}[1]{}
\begin{thebibliography}{38}
\providecommand{\natexlab}[1]{#1}
\providecommand{\url}[1]{\texttt{#1}}
\expandafter\ifx\csname urlstyle\endcsname\relax
  \providecommand{\doi}[1]{doi: #1}\else
  \providecommand{\doi}{doi: \begingroup \urlstyle{rm}\Url}\fi

\bibitem[Xie et~al.(2013)Xie, Singh, Uang, Narayan, and
  Abbeel]{xie2013multimodal}
Z.~Xie, A.~Singh, J.~Uang, K.~S. Narayan, and P.~Abbeel.
\newblock Multimodal blending for high-accuracy instance recognition.
\newblock In \emph{IROS}, pages 2214--2221. IEEE, 2013.

\bibitem[Held et~al.(2016)Held, Thrun, and Savarese]{held2016robust}
D.~Held, S.~Thrun, and S.~Savarese.
\newblock Robust single-view instance recognition.
\newblock In \emph{Robotics and Automation (ICRA), 2016 IEEE International
  Conference on}, pages 2152--2159. IEEE, 2016.

\bibitem[Georgakis et~al.(2016)Georgakis, Reza, Mousavian, Le, and
  Kosecka]{georgakis2016multiview}
G.~Georgakis, M.~A. Reza, A.~Mousavian, P.-H. Le, and J.~Kosecka.
\newblock Multiview rgb-d dataset for object instance detection.
\newblock \emph{arXiv preprint arXiv:1609.07826}, 2016.

\bibitem[Lowe(2004)]{lowe2004distinctive}
D.~G. Lowe.
\newblock Distinctive image features from scale-invariant keypoints.
\newblock \emph{IJCV}, 60\penalty0 (2):\penalty0 91--110, 2004.

\bibitem[Bay et~al.(2006)Bay, Tuytelaars, and Van~Gool]{bay2006surf}
H.~Bay, T.~Tuytelaars, and L.~Van~Gool.
\newblock Surf: Speeded up robust features.
\newblock In \emph{ECCV}, pages 404--417. Springer, 2006.

\bibitem[Quadros et~al.(2012)Quadros, Underwood, and
  Douillard]{quadros2012occlusion}
A.~Quadros, J.~P. Underwood, and B.~Douillard.
\newblock An occlusion-aware feature for range images.
\newblock In \emph{Robotics and Automation (ICRA), 2012 IEEE International
  Conference on}, pages 4428--4435. IEEE, 2012.

\bibitem[Huttenlocher et~al.(1993)Huttenlocher, Klanderman, and
  Rucklidge]{huttenlocher1993comparing}
D.~P. Huttenlocher, G.~A. Klanderman, and W.~J. Rucklidge.
\newblock Comparing images using the hausdorff distance.
\newblock \emph{PAMI}, 15\penalty0 (9):\penalty0 850--863, 1993.

\bibitem[Hinterstoisser et~al.(2012)Hinterstoisser, Cagniart, Ilic, Sturm,
  Navab, Fua, and Lepetit]{hinterstoisser2012gradient}
S.~Hinterstoisser, C.~Cagniart, S.~Ilic, P.~Sturm, N.~Navab, P.~Fua, and
  V.~Lepetit.
\newblock Gradient response maps for real-time detection of textureless
  objects.
\newblock \emph{PAMI}, 34\penalty0 (5):\penalty0 876--888, 2012.

\bibitem[Lai et~al.(2011)Lai, Bo, Ren, and Fox]{lai2011large}
K.~Lai, L.~Bo, X.~Ren, and D.~Fox.
\newblock A large-scale hierarchical multi-view rgb-d object dataset.
\newblock In \emph{Robotics and Automation (ICRA), 2011 IEEE International
  Conference on}, pages 1817--1824. IEEE, 2011.

\bibitem[Georgakis et~al.(2017)Georgakis, Mousavian, Berg, and
  Kosecka]{georgakis2017synthesizing}
G.~Georgakis, A.~Mousavian, A.~C. Berg, and J.~Kosecka.
\newblock Synthesizing training data for object detection in indoor scenes.
\newblock \emph{arXiv preprint arXiv:1702.07836}, 2017.

\bibitem[Ammirato et~al.(2018)Ammirato, Fu, Shvets, Kosecka, and Berg]{tdid}
P.~Ammirato, C.-Y. Fu, M.~Shvets, J.~Kosecka, and A.~C. Berg.
\newblock Target driven instance detection.
\newblock \emph{arXiv preprint arXiv:1803.04610}, 2018.

\bibitem[Koch(2015)]{koch2015siamese}
G.~Koch.
\newblock Siamese neural networks for one-shot image recognition.
\newblock 2015.

\bibitem[Merler et~al.(2007)Merler, Galleguillos, and
  Belongie]{merler2007recognizing}
M.~Merler, C.~Galleguillos, and S.~Belongie.
\newblock Recognizing groceries in situ using in vitro training data.
\newblock In \emph{2007 IEEE Conference on Computer Vision and Pattern
  Recognition}, pages 1--8. IEEE, 2007.

\bibitem[George and Floerkemeier(2014)]{george2014recognizing}
M.~George and C.~Floerkemeier.
\newblock Recognizing products: A per-exemplar multi-label image classification
  approach.
\newblock In \emph{European Conference on Computer Vision}, pages 440--455.
  Springer, 2014.

\bibitem[Franco et~al.(2017)Franco, Maltoni, and Papi]{franco2017grocery}
A.~Franco, D.~Maltoni, and S.~Papi.
\newblock Grocery product detection and recognition.
\newblock \emph{Expert Systems with Applications}, 81:\penalty0 163--176, 2017.

\bibitem[Geng et~al.(2018)Geng, Han, Lin, Zhu, Bai, Wang, He, Xiao, and
  Lai]{geng2018fine}
W.~Geng, F.~Han, J.~Lin, L.~Zhu, J.~Bai, S.~Wang, L.~He, Q.~Xiao, and Z.~Lai.
\newblock Fine-grained grocery product recognition by one-shot learning.
\newblock In \emph{2018 ACM Multimedia Conference on Multimedia Conference},
  pages 1706--1714. ACM, 2018.

\bibitem[Choy et~al.(2016)Choy, Gwak, Savarese, and
  Chandraker]{choy2016universal}
C.~B. Choy, J.~Gwak, S.~Savarese, and M.~Chandraker.
\newblock Universal correspondence network.
\newblock In \emph{Advances in Neural Information Processing Systems}, pages
  2414--2422, 2016.

\bibitem[Florence et~al.(2018)Florence, Manuelli, and
  Tedrake]{florence2018dense}
P.~R. Florence, L.~Manuelli, and R.~Tedrake.
\newblock Dense object nets: Learning dense visual object descriptors by and
  for robotic manipulation.
\newblock \emph{arXiv preprint arXiv:1806.08756}, 2018.

\bibitem[Dosovitskiy et~al.(2015)Dosovitskiy, Fischer, Ilg, Hausser, Hazirbas,
  Golkov, Van Der~Smagt, Cremers, and Brox]{dosovitskiy2015flownet}
A.~Dosovitskiy, P.~Fischer, E.~Ilg, P.~Hausser, C.~Hazirbas, V.~Golkov, P.~Van
  Der~Smagt, D.~Cremers, and T.~Brox.
\newblock Flownet: Learning optical flow with convolutional networks.
\newblock In \emph{Proceedings of the IEEE International Conference on Computer
  Vision}, pages 2758--2766, 2015.

\bibitem[Ilg et~al.(2017)Ilg, Mayer, Saikia, Keuper, Dosovitskiy, and
  Brox]{ilg2017flownet}
E.~Ilg, N.~Mayer, T.~Saikia, M.~Keuper, A.~Dosovitskiy, and T.~Brox.
\newblock Flownet 2.0: Evolution of optical flow estimation with deep networks.
\newblock In \emph{IEEE conference on computer vision and pattern recognition
  (CVPR)}, volume~2, page~6, 2017.

\bibitem[Chow(1970)]{chow1970optimum}
C.~Chow.
\newblock On optimum recognition error and reject tradeoff.
\newblock \emph{IEEE Transactions on information theory}, 16\penalty0
  (1):\penalty0 41--46, 1970.

\bibitem[Bartlett and Wegkamp(2008)]{bartlett2008classification}
P.~L. Bartlett and M.~H. Wegkamp.
\newblock Classification with a reject option using a hinge loss.
\newblock \emph{Journal of Machine Learning Research}, 9\penalty0
  (Aug):\penalty0 1823--1840, 2008.

\bibitem[Cortes et~al.(2016)Cortes, DeSalvo, and Mohri]{cortes2016boosting}
C.~Cortes, G.~DeSalvo, and M.~Mohri.
\newblock Boosting with abstention.
\newblock In \emph{Advances in Neural Information Processing Systems}, pages
  1660--1668, 2016.

\bibitem[Geifman and El-Yaniv(2019)]{geifman2019selectivenet}
Y.~Geifman and R.~El-Yaniv.
\newblock Selectivenet: A deep neural network with an integrated reject option.
\newblock \emph{arXiv preprint arXiv:1901.09192}, 2019.

\bibitem[Cortes et~al.(2016)Cortes, DeSalvo, and Mohri]{cortes2016learning}
C.~Cortes, G.~DeSalvo, and M.~Mohri.
\newblock Learning with rejection.
\newblock In \emph{International Conference on Algorithmic Learning Theory},
  pages 67--82. Springer, 2016.

\bibitem[Shafer and Vovk(2008)]{shafer2008tutorial}
G.~Shafer and V.~Vovk.
\newblock A tutorial on conformal prediction.
\newblock \emph{Journal of Machine Learning Research}, 9\penalty0
  (Mar):\penalty0 371--421, 2008.

\bibitem[Hechtlinger et~al.(2018)Hechtlinger, P{\'o}czos, and
  Wasserman]{hechtlinger2018cautious}
Y.~Hechtlinger, B.~P{\'o}czos, and L.~Wasserman.
\newblock Cautious deep learning.
\newblock \emph{arXiv preprint arXiv:1805.09460}, 2018.

\bibitem[Fischler and Bolles(1981)]{fischler1981random}
M.~A. Fischler and R.~C. Bolles.
\newblock Random sample consensus: a paradigm for model fitting with
  applications to image analysis and automated cartography.
\newblock \emph{Communications of the ACM}, 24\penalty0 (6):\penalty0 381--395,
  1981.

\bibitem[Longuet-Higgins(1981)]{longuet1981computer}
H.~C. Longuet-Higgins.
\newblock A computer algorithm for reconstructing a scene from two projections.
\newblock \emph{Nature}, 293\penalty0 (5828):\penalty0 133, 1981.

\bibitem[Georgakis et~al.(2016)Georgakis, Reza, Mousavian, Le, and
  Kosecka]{gmuk}
G.~Georgakis, M.~A. Reza, A.~Mousavian, P.-H. Le, and J.~Kosecka.
\newblock Multiview rgb-d dataset for object instance detection.
\newblock \emph{arXiv preprint arXiv:1609.07826}, 2016.

\bibitem[Lai et~al.(2014)Lai, Bo, and Fox]{lai2014unsupervised}
K.~Lai, L.~Bo, and D.~Fox.
\newblock Unsupervised feature learning for 3d scene labeling.
\newblock In \emph{Robotics and Automation (ICRA), 2014 IEEE International
  Conference on}, pages 3050--3057. IEEE, 2014.

\bibitem[Singh et~al.(2014)Singh, Sha, Narayan, Achim, and Abbeel]{bigbird}
A.~Singh, J.~Sha, K.~S. Narayan, T.~Achim, and P.~Abbeel.
\newblock Bigbird: A large-scale 3d database of object instances.
\newblock In \emph{Robotics and Automation (ICRA), 2014 IEEE International
  Conference on}, pages 509--516. IEEE, 2014.

\bibitem[{Lai} et~al.(2011){Lai}, {Bo}, {Ren}, and {Fox}]{5980382}
K.~{Lai}, L.~{Bo}, X.~{Ren}, and D.~{Fox}.
\newblock A large-scale hierarchical multi-view rgb-d object dataset.
\newblock In \emph{2011 IEEE International Conference on Robotics and
  Automation}, pages 1817--1824, May 2011.
\newblock \doi{10.1109/ICRA.2011.5980382}.

\bibitem[Ammirato et~al.(2017)Ammirato, Poirson, Park, Ko{\v{s}}eck{\'a}, and
  Berg]{avd}
P.~Ammirato, P.~Poirson, E.~Park, J.~Ko{\v{s}}eck{\'a}, and A.~C. Berg.
\newblock A dataset for developing and benchmarking active vision.
\newblock In \emph{Robotics and Automation (ICRA), 2017 IEEE International
  Conference on}, pages 1378--1385. IEEE, 2017.

\bibitem[Dwibedi et~al.(2017)Dwibedi, Misra, and Hebert]{dwibedi2017cut}
D.~Dwibedi, I.~Misra, and M.~Hebert.
\newblock Cut, paste and learn: Surprisingly easy synthesis for instance
  detection.
\newblock In \emph{The IEEE international conference on computer vision
  (ICCV)}, 2017.

\bibitem[Lin et~al.(2014)Lin, Maire, Belongie, Hays, Perona, Ramanan,
  Doll{\'a}r, and Zitnick]{lin2014microsoft}
T.-Y. Lin, M.~Maire, S.~Belongie, J.~Hays, P.~Perona, D.~Ramanan,
  P.~Doll{\'a}r, and C.~L. Zitnick.
\newblock Microsoft coco: Common objects in context.
\newblock In \emph{European conference on computer vision}, pages 740--755.
  Springer, 2014.

\bibitem[Lowe(1999)]{lowe1999object}
D.~G. Lowe.
\newblock Object recognition from local scale-invariant features.
\newblock In \emph{ICCV}, volume~2, pages 1150--1157. IEEE, 1999.

\bibitem[Liu et~al.(2018)Liu, Lehman, Molino, Such, Frank, Sergeev, and
  Yosinski]{liu2018intriguing}
R.~Liu, J.~Lehman, P.~Molino, F.~P. Such, E.~Frank, A.~Sergeev, and
  J.~Yosinski.
\newblock An intriguing failing of convolutional neural networks and the
  coordconv solution.
\newblock In \emph{Advances in Neural Information Processing Systems}, pages
  9628--9639, 2018.

\end{thebibliography}

\clearpage

\renewcommand{\appendixpagename}{\centering Appendix}
\begin{appendices}
  \section{Theoretical Framework}
\label{supp:theoretical-framework}

In this section, we lay out a theoretical framework for  instance detection with no false positives. We will specify a set of assumptions and prove that our detector will have no false positives if these assumptions are satisfied.  Our practical system will make approximations to these assumptions; nonetheless the below theoretical framework will form the basis to our approach.

\textbf{Notation}. We denote by $O_i$ the $i$th object from the $O$ categories of objects.   For each object $O_i$, we assume access to a dataset of $M_i$ template images recorded from a variety of viewpoints and lighting conditions; we denote $I_{i,m}$ as the $m$th template image for object $O_i$.

We denote by $T : (u_1,v_1) \to (u_2, v_2)$ a 2D mapping from pixels of one image to the pixels of another image.  We overload notation such that $T(I)$ is an application of the 2D mapping $T$ to all pixels in the image $I$ to produce a new image $T(I)$.  We denote by $T^R$ the set of such 2D mappings which are perfectly rigid, i.e. mappings that could be derived from some rigid object transformation.  In other words, if $T \in T^R$, then there exists some Fundamental Matrix $F$ such that, for all $x = (u,v)$, we have that $x'^\top F x = 0$, where $x' = T(u, v)$. Let $r(T)$ denote the ``rigidity" of such a transformation, measured by the fraction of inlier pixel matches under the best approximating rigid transformation.  In other words, for any given transformation $T$, we find 
\begin{align}
    r(T) = \max_{\bar{T} \in T^R} \frac{\sum_{u,v} \mathds{1}\{||T(u, v) - \bar{T}(u, v)|| \leq \epsilon\}}{\sum_{u,v} 1}
\end{align}
where $\mathds{1}\{ \cdot \}$ is the indicator function and $\epsilon$ is a constant.  We describe our practical implementation of this function in Section~\ref{subsec:verification-tests}.  Since $T^R$ is the set of perfectly rigid transformations, for all $T \in T^R$, we have that $r(T) = 1$. Also, if $\epsilon = 0$, $r(T) = 1$ implies that $T \in T^R$. However for practical reasons, we relax $\epsilon$ to a non-zero but small value in our implementation. 

Let $D$ denote a detection in a scene image (e.g. $D$ is a crop of a scene image).  We also denote $gt(D)$ as the ground-truth object class for the object contained in the image $D$. Note that we are implicitly assuming that all detections contain an image of some object in our dataset.

%which contains an image of an object with corresponding ground-truth object class of $O_i$. 

We assume access to a similarity classifier which returns a score $c(I_1, I_2)$ that indicates the confidence that images $I_1$ and $I_2$ each contain the same object class. Practically, this corresponds to our base one-shot instance detector. We will specify below the assumptions that we make about the similarity classifier.

%this could be any distance metric that satisfies certain properties specified in Appendix~\ref{supp:theoretical-framework}, such as  pixelwise $L2$-distance or normalized cross-correlation.

We also assume access to a distance metric $d(I_1, I_2)$ which measures the the distance between images $I_1$ and $I_2$; 
Any distance function $d:\mathcal{I}^2 \rightarrow \mathbb{R}^+$ that satisfies the following two properties can be used (the properties will be necessary for the proofs later):
\begin{enumerate}
    \item Triangle Inequality: $d(I_1, I_2) \leq d(I_1, I_3) + d(I_2, I_3)$.
    \item Permutational Invariance: $d(I_1, I_2) = d(\sigma(I_1), \sigma(I_2))$, where $\sigma$ is any permutation of image pixels.
\end{enumerate}
Note that many distance metrics satisfy the above properties, such as the maximum difference in pixel intensities $d(I_1, I_2) = ||I_1 - I_2||_\infty$, any $L_p$ norm, or normalized cross-correlation that we implement in our system.

We denote $F(I_1, I_2)$ as a function that computes a 2D mapping between images $I_1$ and $I_2$.  Typically, the objective of this function is to find a rigid 2D mapping that minimizes some distance metric $d$, i.e. 
\begin{align}
    \arg \min_{T \in T^R} d(T(I_1), I_2)
    \label{eq:min_flow}
\end{align}
However, in the below theorem, we make no assumptions about the output of $F(I_1, I_2)$; we will prove that our system will have no false positives, regardless of the output of $F(I_1, I_2)$.  Nonetheless, if $F(I_1, I_2)$ does output a transformation that optimizes Equation~\ref{eq:min_flow}, then this will maximize the recall of our algorithm.

%We denote $F(O_{i, m}, D)$ as a function that computes a 2D mapping between target image $I_{i,m}$ and detection $D$.  Typically, the objective of this function is to find a 2D mapping

%. The result is a transformation $T$

%We want the distance metric $d(I_1, I_2)$ to denote a  distance between two images. 

We make the below assumptions; we will prove that, with these assumptions, our algorithm will lead to no false positives.  In practice, we will need to relax these assumptions for a practical implementation; nonetheless, this theoretical framework provides the basis for our approach.
\\

% %\begin{assumption}
%     \textbf{Assumption 1. (Generative Process of Proposed Detections)}\\
%     Every proposed detection $D$ is generated by the following procedure:
%     \begin{enumerate}
%         \item Sample a continuous viewpoint $v$ of some object $O_i$ and render an image at that viewpoint with a fixed lighting to form an image $D_{temp}$%i.e. $D := O_{i,v}$.
%         \item Apply a limited lighting change to the rendered image to create image $D$, where $d(D_{temp}, D) \leq \gamma$.
%     \end{enumerate}
% %\end{assumption}
% where $\gamma$ is a constant. This assumption states that all proposed detections are generated from objects in our dataset from some viewpoint and a bounded amount of lighting variation; this assumption implicitly ignores occlusions or background variation, although our practical implementation handles such cases. 

% %, i.e. $D := D + \epsilon$ where  $d(D', D) \leq \gamma$.

%This essentially states that the detection objects that need to be accepted or rejected, are generated by applying a limited lighting change to a randomly selected viewpoint of the object. It is assumed that there are no occlusions.  

%\begin{assumption}

\textbf{Assumption 1. (Dense Dataset)}\\
For each detection
$D$, which is of class $O_i$, $\exists m \in M_i, T \in T^R$ such that $d(T(I_{i, m}), D) \leq \gamma$.

    %There are enough viewpoints in the training set for each object, such that given $D_i$, $\exists m, T^R$ such that $d(T^R(O_{i, m}), D_i) \leq \gamma$.
%\end{assumption}

This assumption states that our dataset is dense enough such that every detection can be constructed as a rigid transformation of some image in the dataset, with a bounded lighting change applied. Note that this assumption implies that all detections contain an image of some object in our dataset, and implicitly ignores occlusions or background variation, although our practical implementation handles occlusions and background variation for the detections. 

%\textit{Justification}: Under assumption 1 , if the number of viewpoints is high enough, we can assume that for any pose of the test detection $D$, there would exist a viewpoint in the training set with a rigid transformation between the two. But due to lighting conditions, perceptual difference in the images could be up to $\gamma$.

%\begin{assumption}
    \textbf{Assumption 2. (Similarity Classifier Smoothness)}\\
    We have a similarity classifier $c$ with a corresponding constant $\delta$ that satisfies the following property: for any two images $I_1$ and $I_2$, and for any detection $D$, if $\exists T \in T^R$ such that $d(T(I_1), I_2) \leq 2\gamma$, then  $|c(I_1, D) - c(I_2, D)| < \delta$.
    
    % We have a similarity detection system $c$ that satisfies the following property -- for any $i, j, m, n$ and $D$ generated according to assumption 1, if $\exists T^R$ such that $d(T^R(O_{i, m}), I_{j,n}) \leq 2\gamma$, then $|c(O_{i, m}, D) - c(I_{j,n}, D)| < \delta$.
%\end{assumption}

% This assumption says that the similarity classifier $c$ is a smooth function: if two inputs (viewpoints of different objects) are similar, then they will produce similar output (confidence scores for detections) at test time.

This assumption states that the similarity classifier $c$ is a smooth function, in the following sense: if two images $I_1$ and $I_2$ are sufficiently similar such that $I_2$ can be created by a rigid transformation and a small lighting change applied to $I_1$, then the similarity classifier $c$ will output a similar score when comparing $I_1$ and $D$ as when comparing $I_2$ and $D$.  

Note that we make no other assumptions on the output of the similarity classifier $c$, and hence this similarity classifier is not, by itself, capable of creating the strong ``no false positive" results that we will provide below for our overall system; the similarity classifier must be combined with our other tests for verifying dense correspondences.

Our pipeline for object detection proceeds as follows: we assume that an initial detector finds detections in an image and proposes an object class $O_i$ for each detection $D$.  We then pass the detection and its corresponding proposed class to our verification system \textsc{TheoreticalFlowVerify}($i, D$) which returns True if it can verify that  $D$ contains an image of class $O_i$ and False otherwise.  Note that $\textsc{TheoreticalFlowVerify}$ may return False either because $O_i$ is the wrong object class or because the algorithm is simply unable to verify the class of the object.

Our algorithm for \textsc{TheoreticalFlowVerify} is described below.  This algorithm is designed to return False for all false positive detections; in other words, \textsc{TheoreticalFlowVerify}$(i, D)$ will return False if $gt(D) \neq O_i$.  The algorithm proceeds as follows: For any detection $D$ and proposed object class $O_i$, \textsc{TheoreticalFlowVerify} iterates over all images $m \in M_i$ in the dataset for object class $O_i$.  It then compares each template image $I_{i,m}$ to the detection $D$ using \textsc{VerifyMatch}, which either returns True if it can verify that $O_i$ is the correct object class of $D$ and False otherwise.

\begin{algorithm}
    \caption{Theoretical Flow Verifier}\label{alg:euclid}
    \begin{algorithmic}[1]
    \Procedure{VerifyMatch}{$I_{i,m}, D$} \Comment{returns True if verification succeeds}
        \State // Test 1: Similar Object Comparison
        \For{$j \in \mathcal{I} \backslash \{i\}$}
            \For{$n \in M_j$}
                \If{$c(I_{i,m}, D) < c(I_{j,n}, D) + \delta$}
                    \State \textbf{return} False
                \EndIf
            \EndFor
        \EndFor
        \State // Test 2: Color Comparision
        \State $\hat{T} \gets F(I_{i,m}, D)$ \Comment{Estimated flow}
        \If{$d(\hat{T}(I_{i,m}), D) > \gamma$}
            \State \textbf{return} False
        \EndIf
        \State // Test 3: Flow Rigidity
        \If{$r(\hat{T}) < 1$}
            \State \textbf{return} False
        \EndIf
        \State \textbf{return} True
    \EndProcedure

    \Procedure{TheoreticalFlowVerify}{$i, D$} %\Comment{returns True if $D$ is of class $i$; otherwise returns False} %label or \texttt{Unknown}}
%        \For{$i \in \mathcal{I}$}
            \For{$m \in M_i$}
                \If {\textsc{VerifyMatch}($I_{i,m}, D)$}
                    \State \textbf{return True} 
                \EndIf
            \EndFor
        %\EndFor
        \State \textbf{return False}  %\texttt{Unknown}
    \EndProcedure
    \end{algorithmic}
\end{algorithm}

\textsc{VerifyMatch} proceeds as follows: First, for template image $I_{i,m}$, it performs ``Similar Object Comparison": it checks whether there is another object $O_j \neq O_i$ in the dataset, and some template image $n$ of object $O_j$, such that $I_{i,m}$ and $I_{j,n}$ both have a sufficiently similar appearance to $D$ according to the similarity classifier $c$.  In such a case, we cannot verify whether $D$ is an object of class $O_i$ or $O_j$, so the method returns False. 

%As we will see below, this step makes use of Assumption 3 about the smoothness of similarity classifier $c$ as well as Assumption 2 of the dense nature of template images $M_i$ and $M_j$ to ensure that we reject any proposed detections that might be mistakenly classified as another similar-looking class.

%If  $I_{i,m}$ and $I_{j,n}$ are sufficiently similar, then we cannot verify, from image $I_{i,m}$, whether $D$ is an object of class $i$, since $I_{i,m}$ and $I_{j,n}$ are sufficiently similar in appearance that $D$ might alternatively be an object of class $j$.  

Next, \textsc{VerifyMatch} computes a set of dense pixel-wise correspondences  $\hat{T}$ (also referred to as ``Flow") between template image $I_{i,m}$ and the detection $D$.  These correspondences $\hat{T}$ represent a 2D mapping from pixels of $I_{i,m}$ to the pixels of $D$, i.e. $\hat{T} : (u_1,v_1) \to (u_2, v_2)$.  Note that we make no assumptions about $\hat{T}$; if our method for estimating correspondences is not very accurate, then this will reduce the recall of our system but we will still have no false positives, since our system will return False for such examples for all object classes. %, since we will verify and reject detections with poor correspondences.

Finally, $\textsc{VerifyMatch}$ checks whether $\hat{T}$ corresponds to a rigid object transformation and returns False otherwise.

Here we state the main theorem of our approach:

%\begin{theorem}
    \textbf{Theorem 1. (No False Positive Theorem)}\\
    Under assumptions 1 and 2, \textsc{TheoreticalFlowVerify} does not produce any false positives. That is, the following statement always holds: $\textsc{TheoreticalFlowVerify}(i, D)$ returns False whenever $gt(D) \neq O_i$, i.e. whenever  the ground-truth class for detection $D$ is different from class $O_i$
    
    %only if the ground-truth class for detection $D$ is class $i$.  
    
    % \in \{i, \texttt{Unknown}\}$.
%\end{theorem}
%    either returns the true label or returns \texttt{Unknown}. 

Note that $\textsc{TheoreticalFlowVerify}(i, D)$ may sometimes return False even if $i$ is the correct ground-truth class of $D$ if it cannot verify this proposal.  In such cases,  \textsc{TheoreticalFlowVerify} will return false for all object classes, meaning that we cannot verify the category of detection $D$ using our approach.  Still, the benefit of our approach is that, when \textsc{TheoreticalFlowVerify}($i, D)$) returns true, we can be assured that detection $D$ is an object of class $O_i$.

%; it will also return False whenever the ground-truth class of $D$ is not $i$.

\begin{proof}
    We need to show that $\textsc{TheoreticalFlowVerify}(i, D)$ returns False whenever the ground-truth class of object $D$ is not $O_i$. Note that from line 17, we need to prove that for any object class $O_i$ that is not equal to the ground truth class, every template image $I_{i,m}$, for all $m \in M_i$ of object $O_i$ needs to be rejected by $\textsc{VerifyMatch}$. That is, $\textsc{VerifyMatch}(I_{i,m}, D) = \text{False}, \forall m \in M_i$, if $gt(D) \neq O_i$. 
    
    Assume that the ground-truth object class of $D$ is $O_j$.  We can partition $M_i$ into three classes:
    \begin{enumerate}
        \item $M_{i,1} := \{m: \exists T \in T^R \text{ such that } d(T(I_{i,m}), D) \leq \gamma \}$.
        \item $M_{i,2} := \{m \in M_i\ |\ d(\hat{T}(I_{i,m}), D) > \gamma\} \backslash M_{i,1}$.
        \item $M_{i,3} := M_i \backslash (M_{i,1} \cup M_{i,2})$.
    \end{enumerate}
    
%        where $\rfil(O_i, D) = \{m: \exists T \in T^R \text{ such that } d(T(I_{i,m}), D) \leq \gamma \}$. In other words, $\rfil(O_i, D)$ is a set of images of object class $O_i$ that can be described by a rigid transformation and a small lighting change applied to $D$ (of object class $O_j$).
    
    Now we show $\textsc{VerifyMatch}(I_{i,m},D) = \text{False}$, for each of the three cases above.

\begin{enumerate}
\item $\underline{\text{Case 1: } m \in M_{i,1}}$.
\textsc{VerifyMatch} will return False at line 6. \\
%Proposition: $m$ gets filtered at line 7.\\

Since $m \in M_{i,1}$, we know that $\exists T_i \in T^R$ such that $d(T_i(I_{i,m}), D) \leq \gamma \}$.  In other words, $M_{i,1}$ is a set of images of object class $O_i$ that can be described by a rigid transformation and a small lighting change applied to $D$ (which is an object of class $O_j$).  Thus there are two object classes, $O_i$ and $O_j$, that both appear similar to detection $D$, although detection $D$ is an object of class $O_j$.

    From assumption 1, $\exists n \in M_j, T_j \in T^{R}$ such that $d(T_j(I_{j,n}), D) \leq \gamma$. Let us define a new operator $T_{ij} = (T_j)^{-1} \circ T_i$, i.e. first apply $T_i$ and then apply inverse of $T_j$. First, note that, if $T_i \in T^R$ and $T_j \in T^R$, then $T_{ij} \in T^R$ since the composition of these rigid transforms is a rigid transform. Secondly,

    \begin{align*}
    d(T_{ij}(I_{i,m}), I_{j,n}) &= d(T_j \circ T_{ij}(I_{i,m}), T_j(I_{j,n}))\\
    &= d(T_i(I_{i,m}), T_j(I_{j,n}))\\
    &\leq d(T_i(I_{i,m}), D) + d(T_j(I_{j,n}), D)\\
    % &\ \ \ \ \ \ \text{(by triangle inequality of $d$)}\\
    &\leq 2\gamma
    \end{align*}
    
    The first line holds by the permutation invariance property of $d$, since $T_j$ is a permutation of pixels.  The second line holds by definition of $T_{ij}$.  The third line holds by triangle inequality of $d$.  The last line holds because of the definition of $m$ and $T_i$ that $d(T_i(I_{i,m}), D) \leq \gamma$, and by definition of $n$ and $T_j$ that $d(T_j(I_{j,n}), D) \leq \gamma$.
    
    %&\ \ \ \ \ \ \text{seen as a permutation of pixels)}\\

    By assumption 2, since $d(T_{ij}(I_{i,m}), I_{j,n}) \leq 2\gamma$, then $|c(I_{i,m}, D) - c(I_{j,n}, D)| < \delta$.  This then implies that
    \begin{align*}
    &c(I_{i,m}, D) < c(I_{j,n}, D) + \delta\\
    &c(I_{i,m}, D) < \max_n c(I_{j,n}, D) + \delta
    \end{align*}
    Hence, the conditional at line 5 will be true and so $\textsc{VerifyMatch}(I_{i,m}, D)$ will return False at line 6.

\item $\underline{\text{Case 2: } m \in M_{i,2}}$.
\textsc{VerifyMatch} will return False at line 10. \\
%Proposition: $m$ gets filtered at line 7 or 10.\\

    %If $m$ does not get filtered at line 7, since 
    
    By definition of $M_{i,2}$, we know that $d(\hat{T}(I_{i,m}), D) > \gamma$.  Hence the conditional at line 9 will be true, so $\textsc{VerifyMatch}(I_{i,m}, D)$ will return False at line 10.
    
    %$m$ will be filtered at line 10 and 

\item $\underline{\text{Case 3: } m \in M_{i,3}}$.
\textsc{VerifyMatch} will return False at line 13. \\
%Proposition: $m$ gets filtered at line 7 or 10 or 13.\\

If $m \in M_{i,3}$, then by definition, $m \notin M_{i,1}$ and $m \notin M_{i,2}$.  Since $m \notin M_{i,1}$, then $\nexists T \in T^R \text{ such that } d(T(I_{i,m}), D) \leq \gamma$.  Since $m \notin M_{i,2}$, then $d(\hat{T}(I_{i,m}), D) \leq \gamma$.  Hence we know that $\hat{T} \notin T^R$, so $r(\hat{T}) < 1$.  Hence the conditional at line 12 will be true, so $\textsc{VerifyMatch}(I_{i,m}, D)$ will return False at line 13.

%    Let's say $m$ doesn't get filtered at line 7 or 10. That means $m \not\in \rfil(O_i, D)$, and $d(\hat{T}(I_{i,m}), D) \leq \gamma$. Note that this implies $r(\hat{T}) < 1$. This is because since there is no rigid transformation between $I_{i,m}$ and $D$ that is $\gamma$-color preserving, but $\hat{T}$ is $\gamma$-color preserving, so $\hat{T}$ cannot be rigid. Then, the conditional in line 12 becomes true. Hence $m$ gets filtered in line 13 and $\textsc{VerifyMatch}(I_{i,m}, D) = \text{False}$.
    \end{enumerate}

Hence we have shown that $\forall i \neq j, \forall m \in M_i, \textsc{VerifyMatch}(I_{i,m}, D) = \text{False}$.\\
This implies that $\textsc{TheoreticalFlowVerify}(i, D) = $ False whenever $i \neq j$. %\in \{j, \texttt{Unknown}\}$.
\end{proof}

  \section{Network Architecture Details}
\label{sec:approach-flow}

% \subsection{Dense Pixel-wise Correspondences}
   % Our method for computing dense pixel-wise correspondeces takes as input two images -- \textit{image1} and \textit{image2}, and computes the correspondence from image1 to image2.
   % To achieve this, we train a deep neural network using a modification of the Flownet \cite{dosovitskiy2015flownet} architecture.
   % FlowNet is typically used to compute the correspondence between two subsequent frames in a video seuqence; however, in our case we need a pixelwise mapping between a labeled training image of a target object and a proposed detection of that object in a cluttered scene.

%   \sid{talk about how flow is usually trained for videos, small displacements in consecutive frames, but we are doing something very different.}
   % Flownet forms the basis of Flownet-2 \cite{ilg2017flownet}, which is a popular optical flow model that has been shown to generalize well to novel scenes.

   We call our network for predicting dense pixel-wise correspondences \flowmatchnet.
For every proposed detection, we pad the detection to square, crop it (which we refer to as the `cropped scene image'), and feed it to \flowmatchnet.
\flowmatchnet\ also takes an image from the dataset and computes a mapping from every pixel in the template image to some pixel in the cropped scene image.
   % Examples of such template and scene images can be found in Figure~\ref{fig:template_and_scene}.

%   \sid{say somewhere that we are using masks for target images}

%    \subsubsection{Architecture}
%FlowNet and other state-of-the-art optical flow models \cite{ilg2017flownet,sun2018pwc} use convolutional neural network architectures.

   We compute pixel correspondence from each template image of the object to the cropped scene image.
   We train a deep neural network using a modification of the FlowNetC neural network architecture~\cite{dosovitskiy2015flownet}.
    Specifically for FlowNetC, a series of convolutional layers are applied separately to input images to extract features from each image.
   Then, a \textit{cross-correlation} layer is used to combine features coming from each image branch, to compare feature in every location in image1 with every location in image2.
   However, FlowNetC was designed for optical flow, which is typically applied to consecutive frames of a video with the assumption that pixels have small displacements between consecutive frames.
   Therefore, in the original FlowNetC, cross-correlation is approximated by considering only a $21\times21$ pixel neighborhood window~\cite{dosovitskiy2015flownet,ilg2017flownet}.
   %Secondly (2), videos are usually of high-resolution, and a full cross-correlation is computationally infeasible.

   However, we cannot make such a `small displacement' assumption in our case because the object in the scene crop might have undergone a fairly large rotation and translation relative to the template image.
   Hence, we perform a full cross correlation between all pixel pairs in the scene and template images.
   Furthermore, flow prediction is traditionally made in terms of displacement vectors.
   Since a full cross-correlations is implemented as unrolling the entire 2D feature map, we reparameterize flow in terms of global coordinates of pixels being mapped to in the second image.
   Hence we concatenate $(x, y)$ pixel coordinates as additional feature maps, following \cite{liu2018intriguing}.
  \section{Training details of \flowmatchnet}
\label{supp:training-flowmatchnet}

\begin{figure}
    \centering
    \newcommand{\scale}{0.3}
    \includegraphics[scale=\scale]{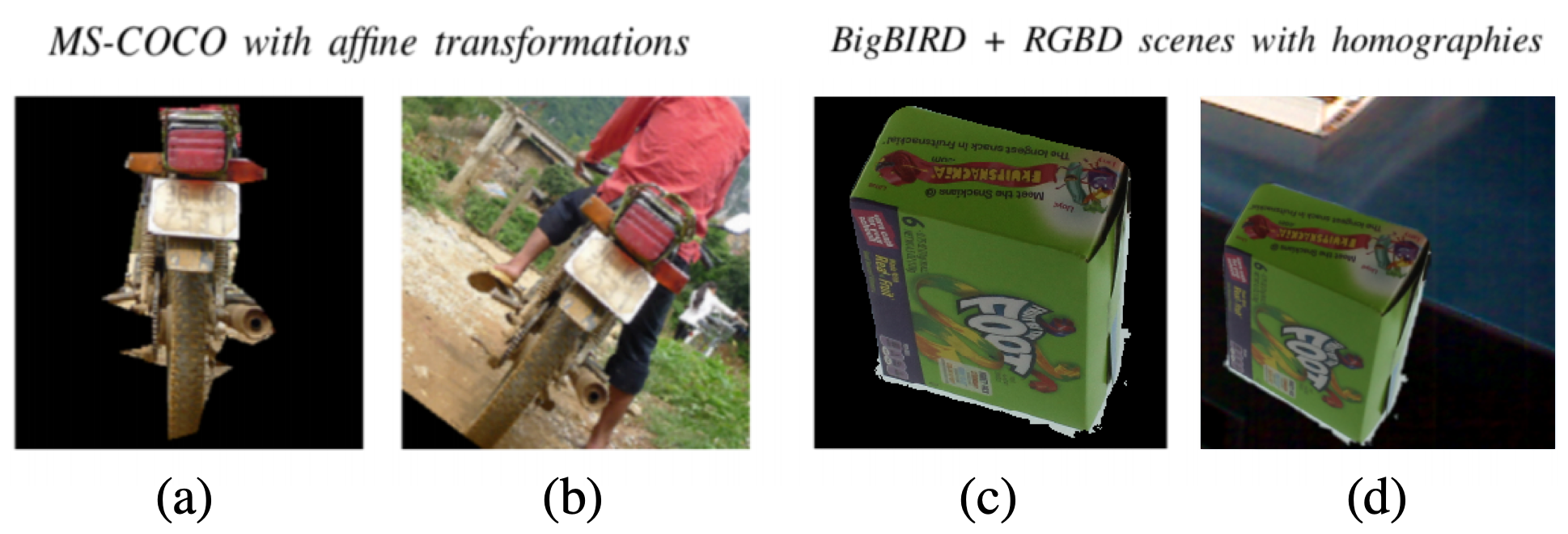}
    \caption{(a), (b): Target and scene images from MS-COCO synthetic dataset that applies affine transformations. (c), (d): Target and scene images from BigBIRD-RGBD synthetic dataset that applies homography transformations.}
    \label{fig:flow-data-augmentation}
\end{figure}

Optical flow models are trained using synthetic datasets that have ground truth flow~\cite{dosovitskiy2015flownet,ilg2017flownet}. Flownet based models have been shown to generalize well when trained on such synthetic datasets.

For training \flowmatchnet, we train our model successively on synthetic datasets with increasing difficulty --

\begin{enumerate}
   \item \emph{MS-COCO with affine transformations}: we take images from the MS-COCO dataset \cite{lin2014microsoft}, crop an area around the object using its annotated bounding box to simualate detections from a base detector, and apply a random rotation and translation to create a simulated scene image.
   These transforations help the flow model to learn simple affine tranformations.
   Although the transformations are relatively simple, MS-COCO has a huge variety of objects; training on such a diverse set of objects can help the network to generalize to different objects types.  Examples of transformations are in Figure~\ref{fig:flow-data-augmentation}.
   \item \emph{BigBIRD + RGBD scenes with homographies}: We take objects from the BigBIRD dataset \cite{bigbird}, crop them out using segmentation masks, and paste them onto background scenes in the W-RGBD scenes dataset \cite{lai2014unsupervised}.
   Since we want \flowmatchnet\ to generalize to novel objects, we exclude training on the 11 objects that are part of GMU Kitchens~\cite{gmuk}, our test set.
   We apply random homographies to the cropped images and blend them into scenes using Cut-Paste-Learn \cite{dwibedi2017cut}, while simulating randomized lighting conditions and image blurs.
   We limit the random homography to ensure that it is not too unrealistic and does not distort the object too much; specifically we ensure that the top-left corner of the bounding box is still the top-left corner after applying the homograph, the bottom-right corner is still at the bottom-right, and so on.
   Importantly, since these are controlled synthetic transformations, ground truth flow can be computed.
   We train on L2 loss for optical flow similar to \cite{dosovitskiy2015flownet,ilg2017flownet}.
\end{enumerate}

We first trained on the MS-COCO synthetic flow dataset for 1.6M iterations using Adam optimizer with a learning rate of $10^{-4}$ with a batch size of 2. We then finetuned on the BigBIRD-RGBD synthetic flow dataset for 50,500 iterations using Adam optimizer with a learning rate of $10^{-6}$ and batch size 2.
  \section{Ablation Details}
\label{supp:ablation}
This section lists detailed results of our ablation analysis in terms of mAP and maximum precision for \flowverify, versus dropping each verification test one at a time. As shown in Table \ref{wrap-tab:new-ablation}, \flowverify\ has the best performance in terms of maximum precision and mAP on GMU-Test, as well the second highest maximum precision on W-RGBD. Dropping \finlier\ results in the largest drop in maximum precision and mAP on both GMU-Test and W-RGBD. This confirms the importance of \textit{flow rigidity} test as suggested by our theoretical framework.

We also observe that removing the \fcolor\ test leads to very little change in performance.  We believe that this is because \flowverify\ is approximately optimizing Supplementary Equation~\ref{eq:min_flow}, with a priority on matching colors:
\begin{align}
    F(I_{i,m}, D)  = \hat{T} \approx \arg \min_{T} d(T(I_{i,m}), D).
    \label{eq:min_flow2}
\end{align}
In other words, \flowverify = $F(I_{i,m}, D)$ returns a transform $\hat{T}$ that tries to match colors in $I_{i,m}$ with similar colors in $D$, thereby minimizing $d(\hat{T}(I_{i,m}), D)$.  Because the \fcolor\ test rejects the detection if $d(\hat{T}(I_{i,m}), D) > \gamma$, our system will very rarely reject a detection based on this criteria.

Additionally, we observe that dropping the \nms\ test on the W-RGBD dataset leads to improvement in mAP of 0.028. We think this is because we set $\eta_{iou}=0.5$ as since it is used for mAP evaluation. However, the optimal value of $\eta_{iou}$ actually depends on how cluttered the objects are in the detection dataset. In practice, if users have some prior knowledge about how cluttered the dataset is, they can adjust $\eta_{iou}$ accordingly and get better performance.
        % \State $\hat{T} \gets F(I_{i,m}, D)$ \If{$d(\hat{T}(I_{i,m}), D) > \gamma$}

%minimizes some distance metric $d$, i.e. normalized cross-correlataion. 

%as a function that computes a 2D mapping between images $I_1$ and $I_2$.  Typically, the objective of this function is to find a rigid 2D mapping that minimizes some distance metric $d$, i.e. 

% \begin{table}[htbp]
%     \makebox[\textwidth][c]{
%     \renewcommand{\arraystretch}{1.1}
%     \begin{tabular}{c|cc|cc}
%     \toprule 
%     & \multicolumn{2}{c|}{GMU-Test} & \multicolumn{2}{c}{RGBD} \\\hline
%     Method & mAP & max Precision  & mAP & max Precision\\\hline
%     %1.41, 1.14, 0.715, 0.357
%     % Baseline                & 0.351    & 0.683 & -    & - \\
%     % TDID-AVD            & - & - & 0.214    & 0.584\\
%     % \hline

%     \flowverify              & 0.401   & \textbf{0.939} & 0.275 & 0.817 \\  \hline
%     \flowverify-\nms         & {\color{red}0.353}    & 0.902 & \textbf{0.306} & 0.730\\  \hline
%     \flowverify-\finlier     & 0.389    & {\color{red}0.808} & {\color{red}0.233} & {\color{red}0.699}\\  \hline
%     \flowverify-\fcolor      & \textbf{0.402}    & 0.933 & 0.275 & \textbf{0.819}\\  \hline
%     \flowverify-\fp          & 0.394    & 0.913 & 0.277 & 0.811\\  \hline
%     \flowverify-\fr          & 0.394    & 0.887 & 0.273 & 0.753\\
%     \bottomrule
%     \end{tabular}
%     \renewcommand{\arraystretch}{1}
%     }\\
%     \caption{Ablation analysis.}
%     \label{wrap-tab:new-ablation}
% \end{table}

\begin{table}[htbp]
    \makebox[\textwidth][c]{
    \renewcommand{\arraystretch}{1.1}
    \begin{tabular}{c|cc|cc}
    \toprule 
    & \multicolumn{2}{c|}{GMU-Test} & \multicolumn{2}{c}{W-RGBD} \\\hline
    Method & mAP & max Precision  & mAP & max Precision\\\hline
    %1.41, 1.14, 0.715, 0.357
    % Baseline                & 0.351    & 0.683 & -    & - \\
    % TDID-AVD            & - & - & 0.214    & 0.584\\
    % \hline

    \flowverify              & \textbf{0.422}   & \textbf{0.939} & 0.278 & 0.822 \\  \hline
    \flowverify-\nms         & 0.412    & 0.906 & \textbf{0.316} & 0.732\\  \hline
    \flowverify-\finlier     & {\color{red}0.394}    & {\color{red}0.807} & {\color{red}0.235} & {\color{red}0.700}\\  \hline
    \flowverify-\fcolor      & \textbf{0.422}    & 0.933 & 0.278 & \textbf{0.823}\\  \hline
    \flowverify-\fp          & 0.415    & 0.913 & 0.280 & 0.811\\  \hline
    \flowverify-\fr          & 0.413    & 0.887 & 0.277 & 0.755\\
    \bottomrule
    \end{tabular}
    \renewcommand{\arraystretch}{1}
    }\\
    \caption{Ablation analysis.}
    \label{wrap-tab:new-ablation}
\end{table}

% (0.389,0.808)
% (0.233,0.699)
% (0.402,0.933)
% (0.275,0.819)
% (0.394,0.913)
% (0.277,0.811)
% (0.394,0.887)
% (0.273,0.753)

  \section{\siftverify}
\label{sec:SiftVerify}
Here we describe the details of the verification tests used for \siftverify. %\textsc{SPrecision}
For \textsc{FPrecision}, we computes the proportion of matched pixels that lie inside the predicted bounding box. This test is often imprecise because SIFT matches are sparse, often as few as 2-3 matches. However, to estimate rigidity using the fundamental matrix, we need at least 8 matches. Hence, instead we simply compute the raw number of SIFT keypoint matches. Since SIFT matches are few in number, it is also not possible to estimate recall the way we do for \fr, so we omit that test as well. We find that these settings give the best performance for this baseline. We run grid-search to find the best thresholds for \siftverify\ on YCB following the same procedure as for \flowverify. 

%\textsc{SPrecision} and \textsc{SMatches} 

%we use \textsc{SMatches} instead, which is 

\section{Further Qualitative Analysis}

In this supplement, we perform a comparative and qualitative analysis of our method \flowverify\ with the \siftverify\ baseline, to complement the quantitative analysis in the main paper. We will first show examples where \siftverify\ succeeds in filtering out false positives from a base detector and where it is successful in retaining true positives. Then we will analyze some failure cases of \siftverify, where our model is able to filter false positives and retain correct detections whereas \siftverify\ is not.

It is important to note that since we are interested in high-precision detection, we will analyze detections which were given a very high confidence score ($>0.99$) by the original detector on the GMU Kitchens \cite{gmuk} test set. In order to improve detector performance in the high precision regime (the leftmost parts of the PR-curve), the system must be able to filter out high-confidence false positives whilst not rejecting too many true detections with high confidence.

As a reminder, \flowverify\ uses five tests -- \nms, \fcolor, \finlier, \fp\ and \fr\ . In order to pass a test, there must exist at least one viewpoint of the target object whose \nms, \fcolor, \finlier, \fp\ and \fr\ scores are all above their respective thresholds. The thresholds that were tuned on YCB-Video train for \fcolor, \finlier, \fp\ and \fr\ turn out to be $0.5$, $0.9$, $0.9$, and $0.3$ respectively, with parameters for \nms\ being $\eta_{iou}=0.5$, $\eta_{diff}=0.0$.

Similarly, SIFT filters detections by computing the number of keypoint matches between the target image and cropped scene image. If the \textsc{SMatches} and \textsc{SPrecision} are above their corresponding thresholds, the detection is deemed to be true by the SIFT baseline. When tuned on YCB-Video Train, the thresholds for \textsc{SMatches} and \textsc{SPrecision} turns out to be $30$ and $0.9$ respectively.

\subsection{Visualization Format}
The visualization format is consistent across all Figure \ref{fig:sfp}, Figure \ref{fig:stp}, Figure \ref{fig:ffp}, and Figure \ref{fig:ftp} -- the top image shows a visualization of matching by \flowmatchnet\ and filtering by \flowverify, whereas the bottom image shows matching using \siftverify. Each visualization consists of two images. On the left is a `target image' -- it is an image of the object being detected. If the detection passes the \flowverify\ or SIFT verification tests, the displayed viewpoint is the one that passes all tests and has the largest product of scores, which denotes the ``best'' viewpoint of the object that is able to be matched by each of the methods. Otherwise, the canonical viewpoint of the object is displayed in the visualization.

%, in one of the 15 viewpoints

At the top of each visualization are scores for tests associated with that method, as well as thresholds, in a \textit{score/threshold} format. 20 randomly-chosen pixel-wise mappings are depicted. In the case of SIFT, sometimes there are less than 20 keypoint matches found. In such cases, all keypoint matches are depicted.

\subsection{SIFT Successful Cases}
In this section, we analyze cases where SIFT matching successfully filters false positives while retaining true detections.

\subsubsection{False Positives rejected by SIFT and \flowverify}
Figure \ref{fig:sfp} contains  four examples that denote false detections that are successfully filtered out by both methods. In these examples , the homography based on flow computed by \flowmatchnet\ produces many outliers, and hence are rejected by the \finlier\ test. This happens because when the object are distinct and do not match, in which case the predicted flow can be arbitrary. SIFT also successfully rejects all these examples, as it produces very few ($<30$) keypoint matches.

\subsubsection{True Positives accepted by SIFT and \flowverify}
Figure \ref{fig:stp} contains fours examples that represent true detections successfully retained by both methods. In these examples, \flowmatchnet and SIFT matches seem nearly perfect.
This section demonstrates that SIFT features can be effective in object matching, and our implementation of SIFT is a reasonably strong baseline.  However, we show below that SIFT can also fail on many cases.

\subsection{SIFT Unsuccessful Cases}
In this section, we analyze cases where SIFT matching fails to filter false positives or retain high-confidence true detections. We illustrate how \flowverify\ successfully handles these cases, to highlight the strengths of our method over the SIFT baseline.

\subsubsection{False Positives accepted by SIFT but rejected by \flowverify}
 In Figure \ref{fig:ffp}, the four examples represent false detections that SIFT fails to filter out, but are successfully filtered out our method.
 In Figure \ref{fig:ffp}(a), the target object is different from the object in the bounding box, but they have the same text in their logos. SIFT fails in this case since it can find enough keypoint matches just in the logo area. \flowverify\ handles this case successfully since it relies on dense pixelwise correspondence. As the object in the cropped scene is different from that in the target image, \flowmatchnet\ cannot find a rigid transformation that matches color for all pixels, thereby not passing the \finlier\ test. 

Figure \ref{fig:ffp}(b, c, d) illustrate another difference between dense and sparse pixel-wise matching. In these examples, the bounding boxes are incomplete and the cropped scene only contains a part of the target object. As SIFT only needs to match a few keypoints, even if small parts of two objects seem to match, it is still enough for SIFT to accept it as a correct detection. \flowverify\ combined with \flowmatchnet\ successfully rejects these as false detections since \flowmatchnet\ computes dense matches across the entire object, resulting in flow fields that have a low \finlier\ score. 

%There are cases where the target is only partially visible in the scene (either partially out of the camera or occluded by other objects). \flowverify\ might introduce false negatives in these scenarios, but it is generally acceptable as we focus more on high-precision regime. 

% Figure \ref{fig:ffp}(d) is another example of an incomplete bounding box. Note that this example is a false positive since it has low IoU ($<$ 0.5) with the ground truth box. SIFT fails in this case as it only matches sparse keypoints. As the box includes the most textured part of the object, SIFT precision is high since most of the keypoint matches are found inside the box. \flowverify\ handles this case through \fp , as a part of pixels are mapped to areas outside the predicted bounding box.

\subsubsection{True Positives rejected by SIFT but retained by \flowverify}
Figure \ref{fig:ftp} contains four examples that denote true positives, which are successfully retained by \flowverify\ but are incorrectly \textit{rejected} by \siftverify.
In all examples (a, b, c, d) in Figure \ref{fig:ftp}, \flowmatchnet\ produces good-quality pixel-wise correspondence, leading to high \fcolor, \finlier, \fp\ and \fr\ scores. However, for most examples, SIFT can produce very few keypoint matches. 

%SIFT has been successfully used to perform feature matching under controlled conditions, such as grocery product detection. However, 

Our experiments show that SIFT struggles to perform reliable keypoint matching in our setting where there can be drastic changes in viewpoints, lighting conditions and even occlusions. We show that in such scenarios, predicting a dense pixel-wise correspondence and designing subsequent verification tests can improve instance detectors towards high-precision and verifiable recognition.

\newcommand{\scale}{0.4}
\begin{figure*}
   \centering
    \begin{subfigure}[t]{\scale\linewidth}
    \includegraphics[width=\textwidth]{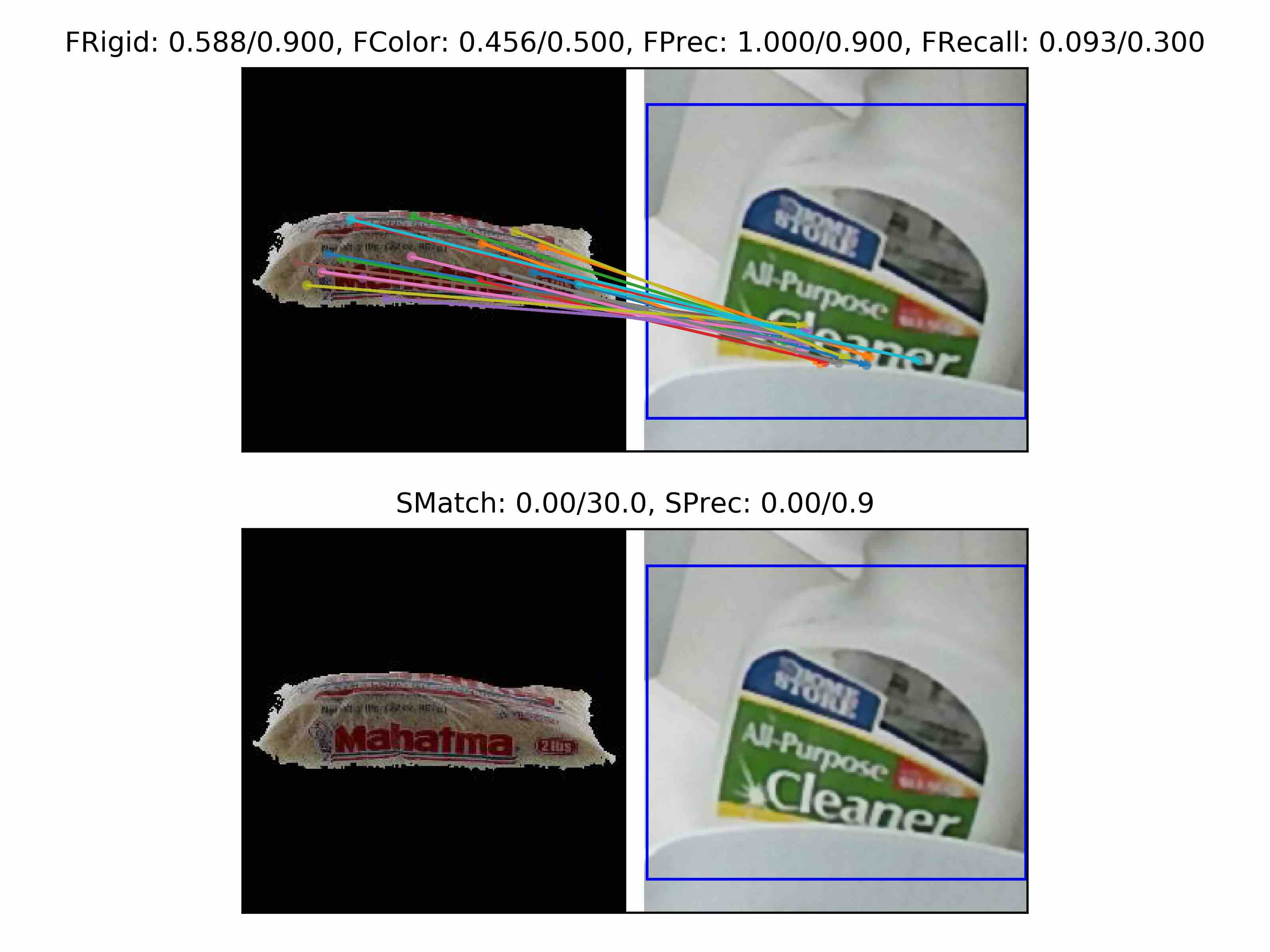}
    \caption{}
    \end{subfigure}
    \begin{subfigure}[t]{\scale\linewidth}
    \includegraphics[width=\textwidth]{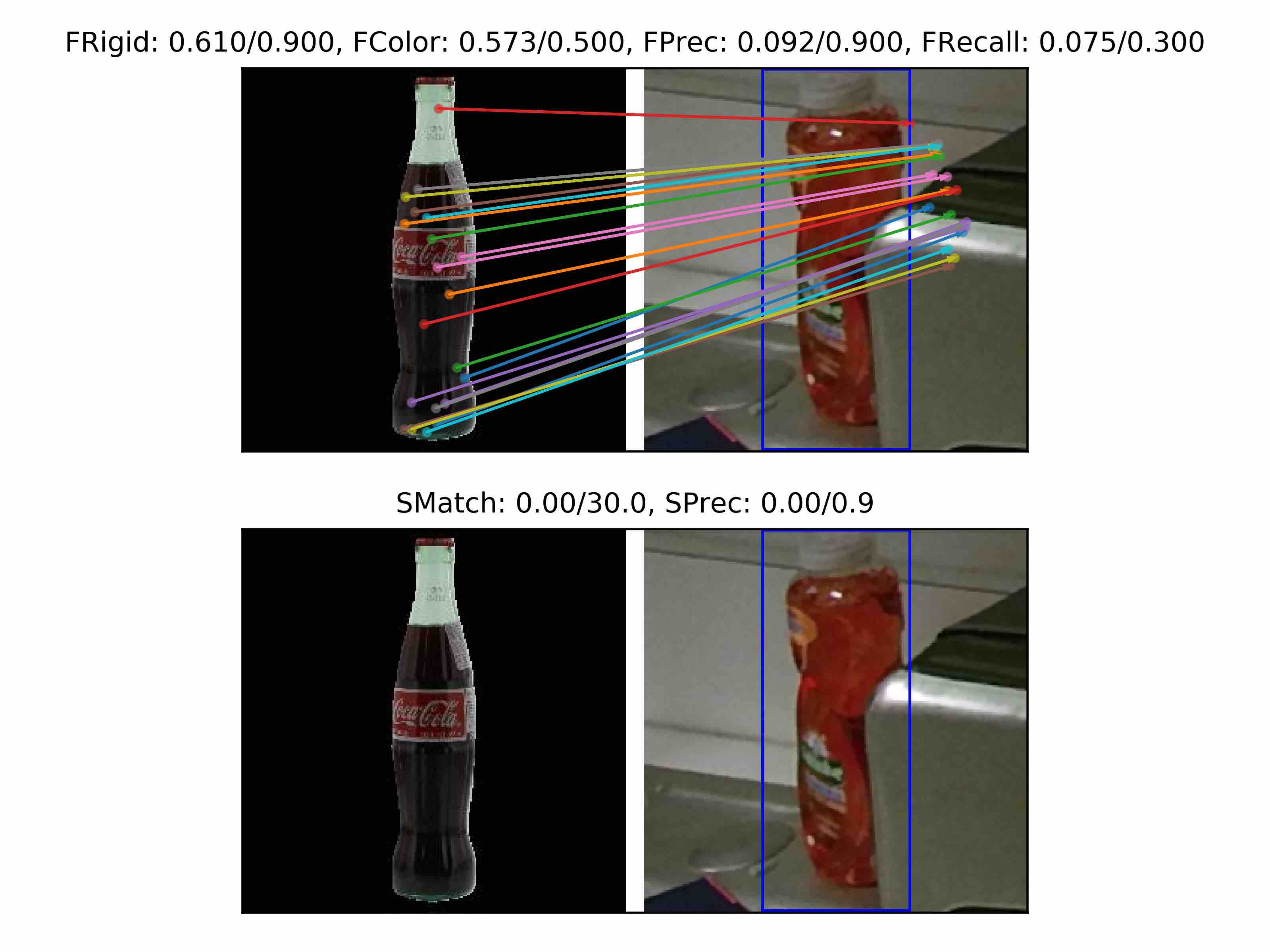}
    \caption{}
    \end{subfigure}\\
    \begin{subfigure}[t]{\scale\linewidth}
    \includegraphics[width=\textwidth]{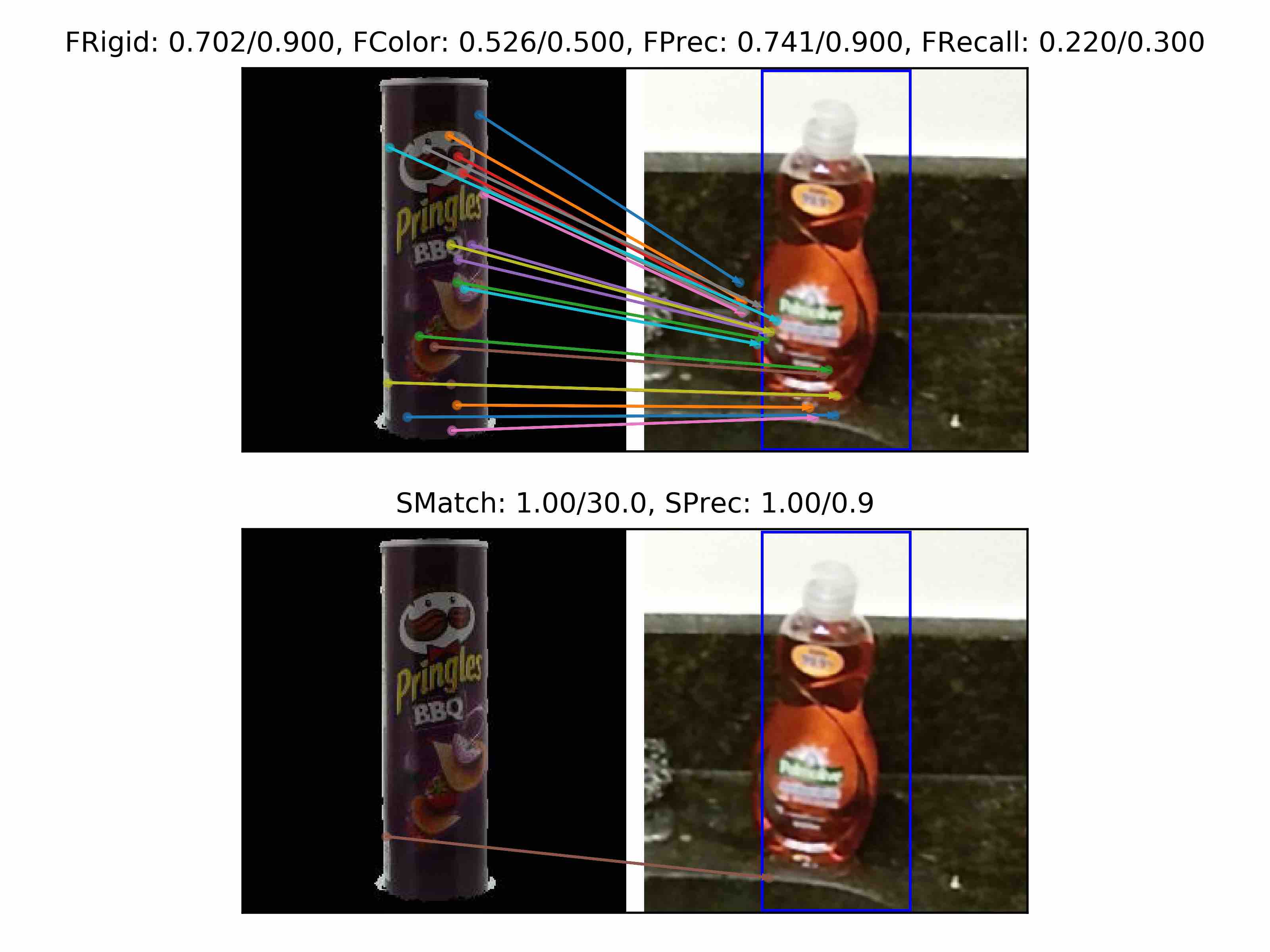}
    \caption{}
    \end{subfigure}
    \begin{subfigure}[t]{\scale\linewidth}
    \includegraphics[width=\textwidth]{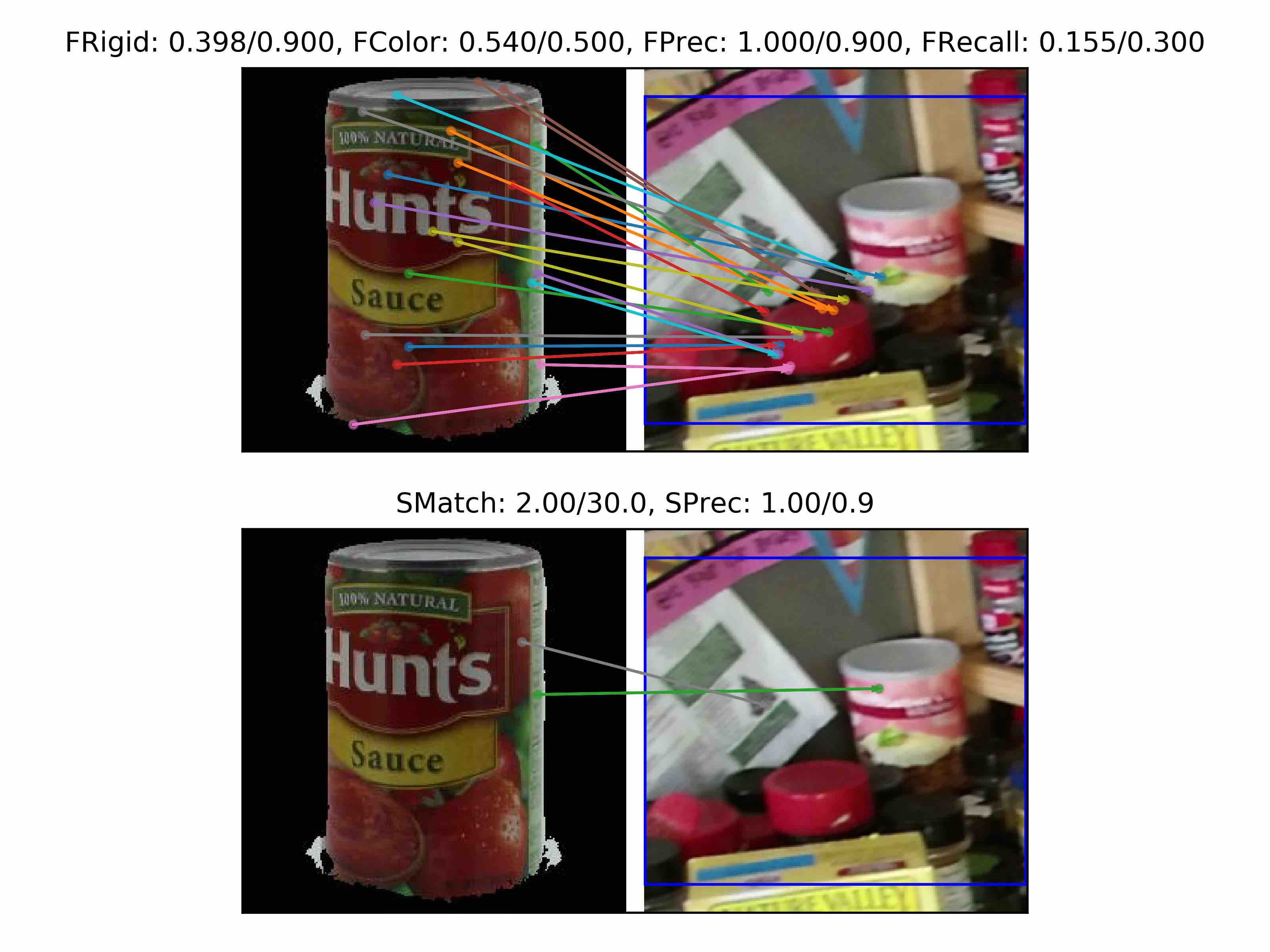}
    \caption{}
    \end{subfigure}
   \caption{False detections which are successfully filtered out by both \flowverify\ and \siftverify. In each image, top: \flowverify, bottom: \siftverify.}
   \label{fig:sfp}
%   \vspace{-30pt}
\end{figure*}

\begin{figure*}
  \centering
    \begin{subfigure}[t]{\scale\linewidth}
    \includegraphics[width=\textwidth]{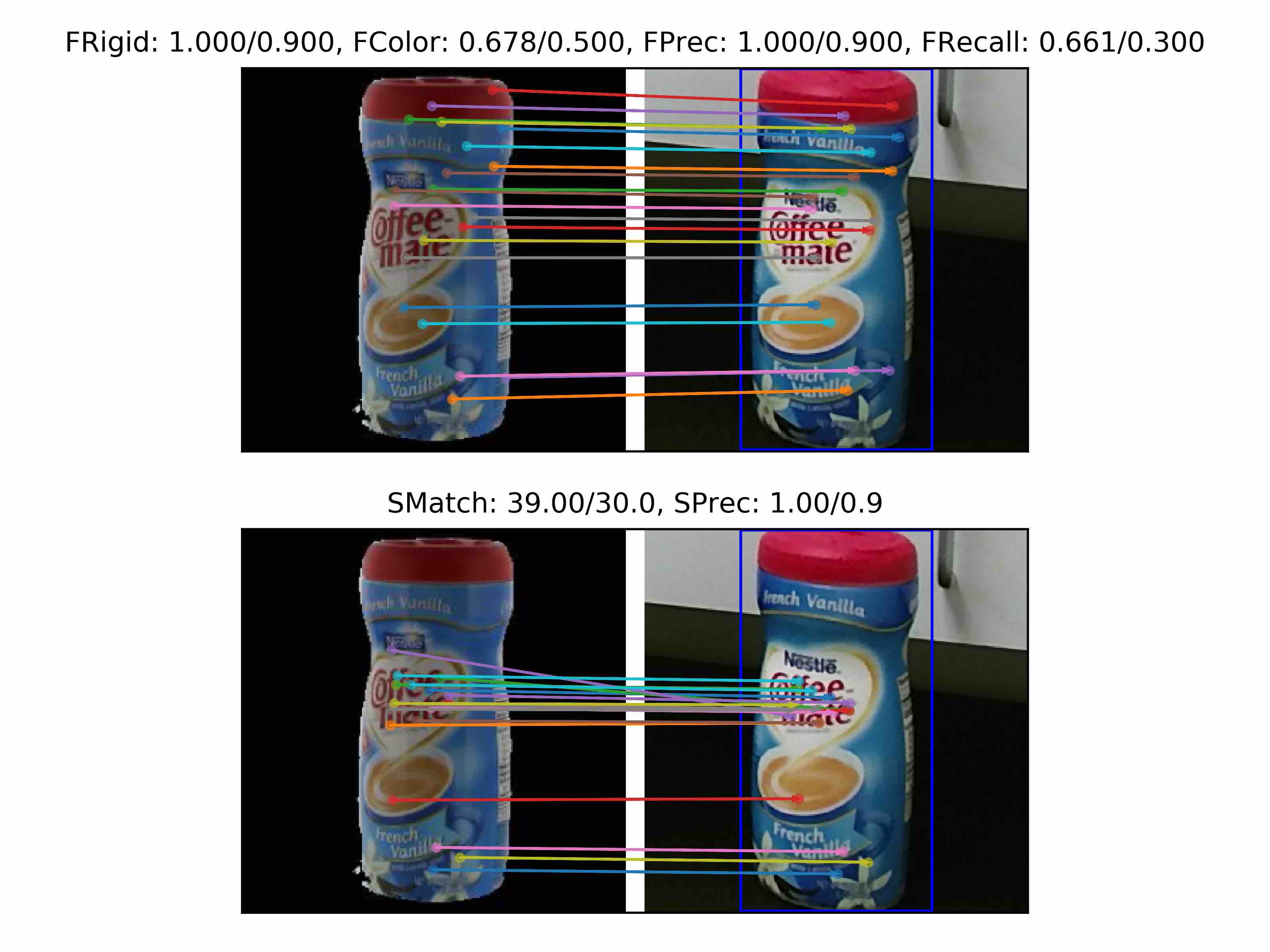}
    \caption{}
    \end{subfigure}
    \begin{subfigure}[t]{\scale\linewidth}
    \includegraphics[width=\textwidth]{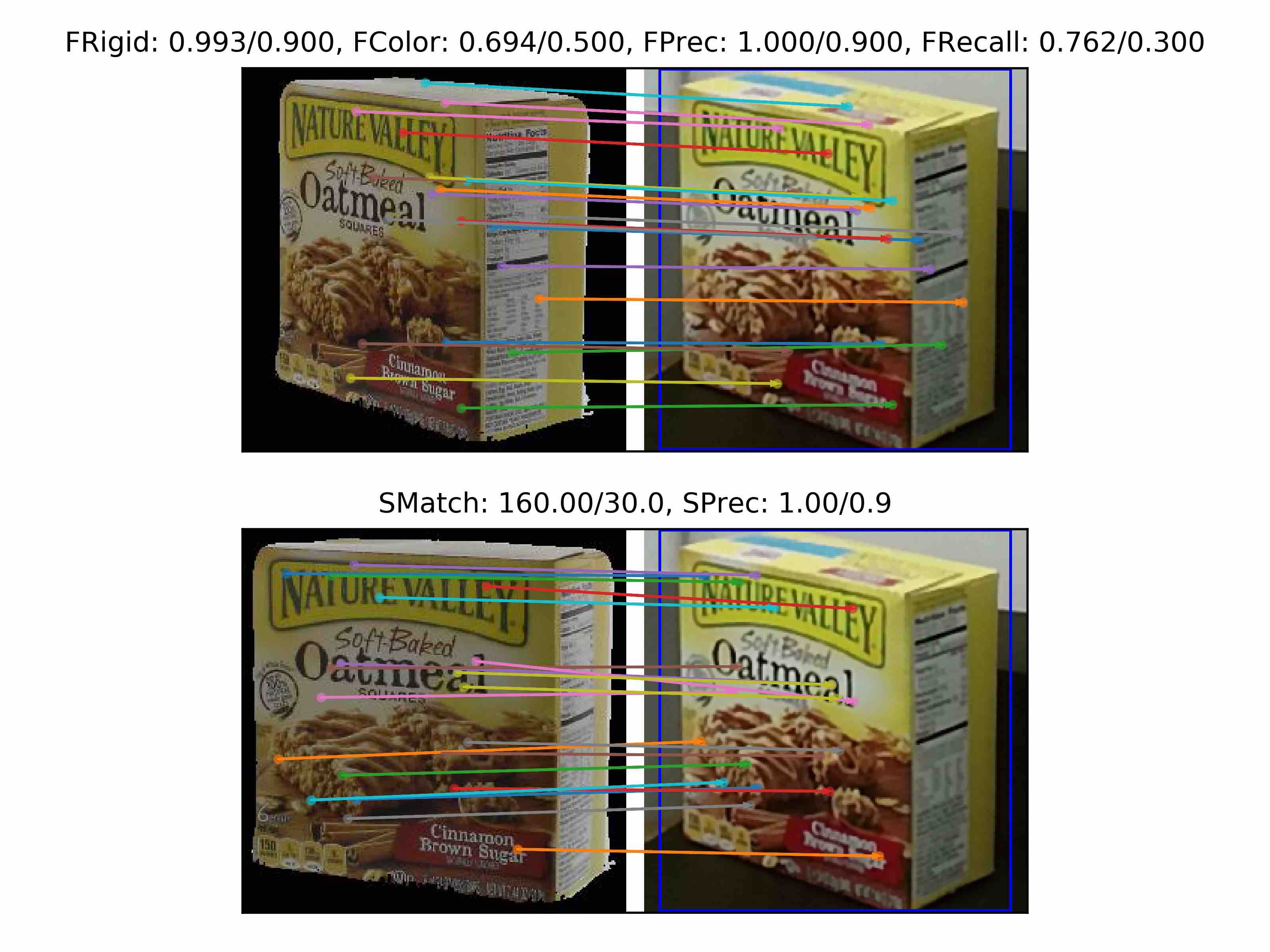}
    \caption{}
    \end{subfigure}
    \begin{subfigure}[t]{\scale\linewidth}
    \includegraphics[width=\textwidth]{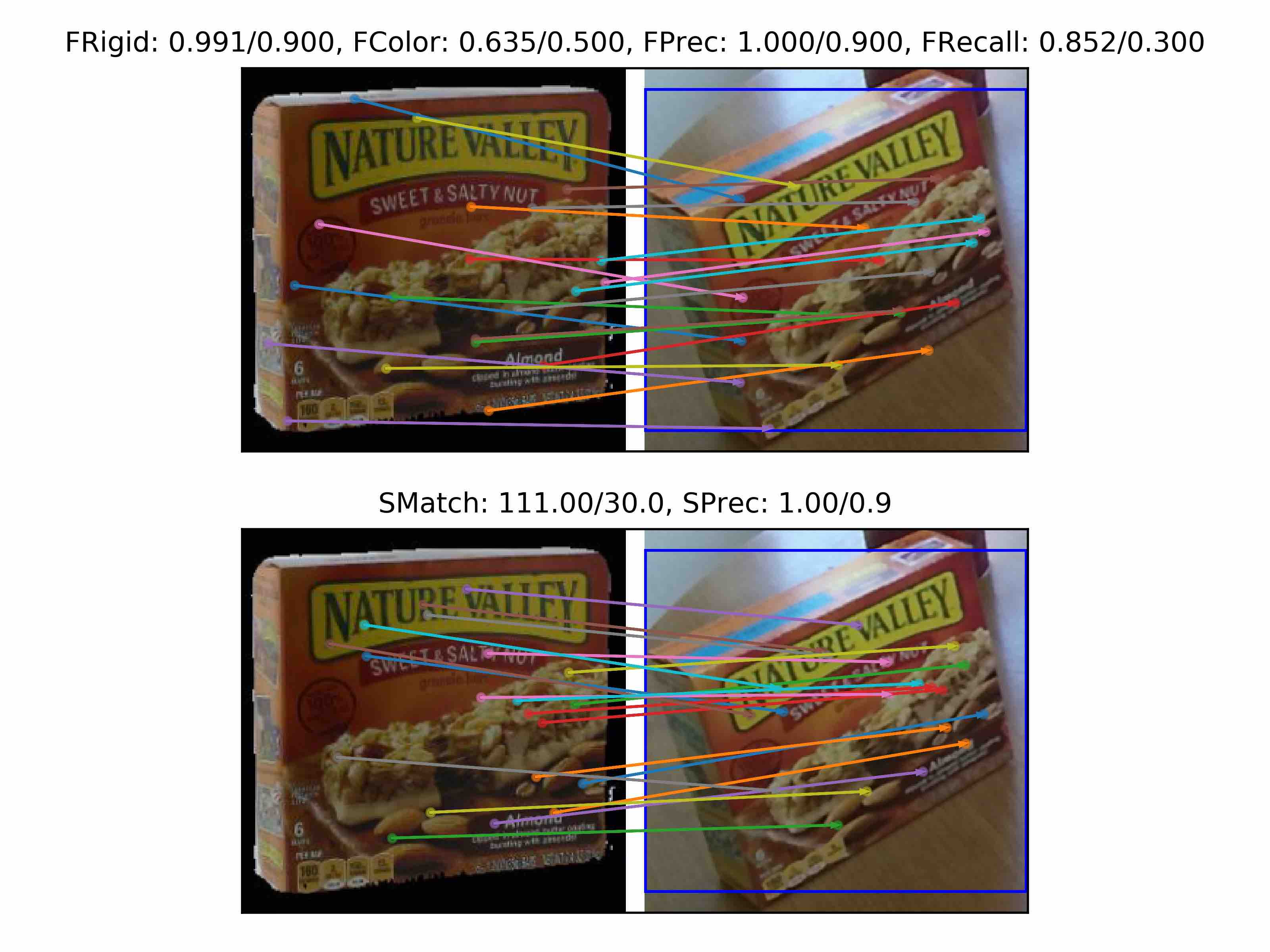}
    \caption{}
    \end{subfigure}
    \begin{subfigure}[t]{\scale\linewidth}
    \includegraphics[width=\textwidth]{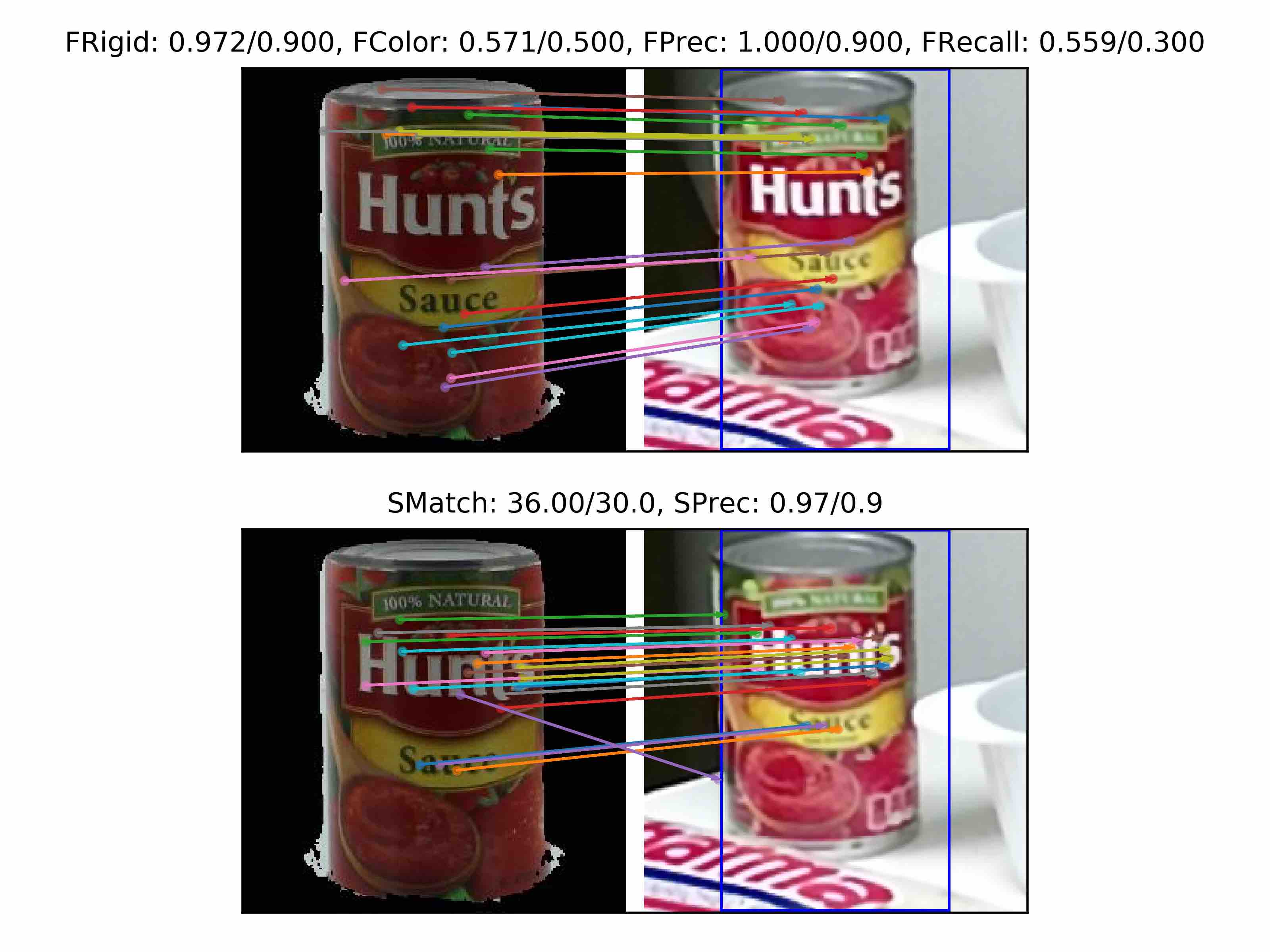}
    \caption{}
    \end{subfigure}
    % \vspace{00pt}
  \caption{True detections that are successfully retained by both \flowverify\ and \siftverify. In each image, top: \flowverify, bottom: \siftverify.}
  \label{fig:stp}
\end{figure*}

\begin{figure*}
    \centering
    \begin{subfigure}[t]{\scale\linewidth}
    \includegraphics[width=\textwidth]{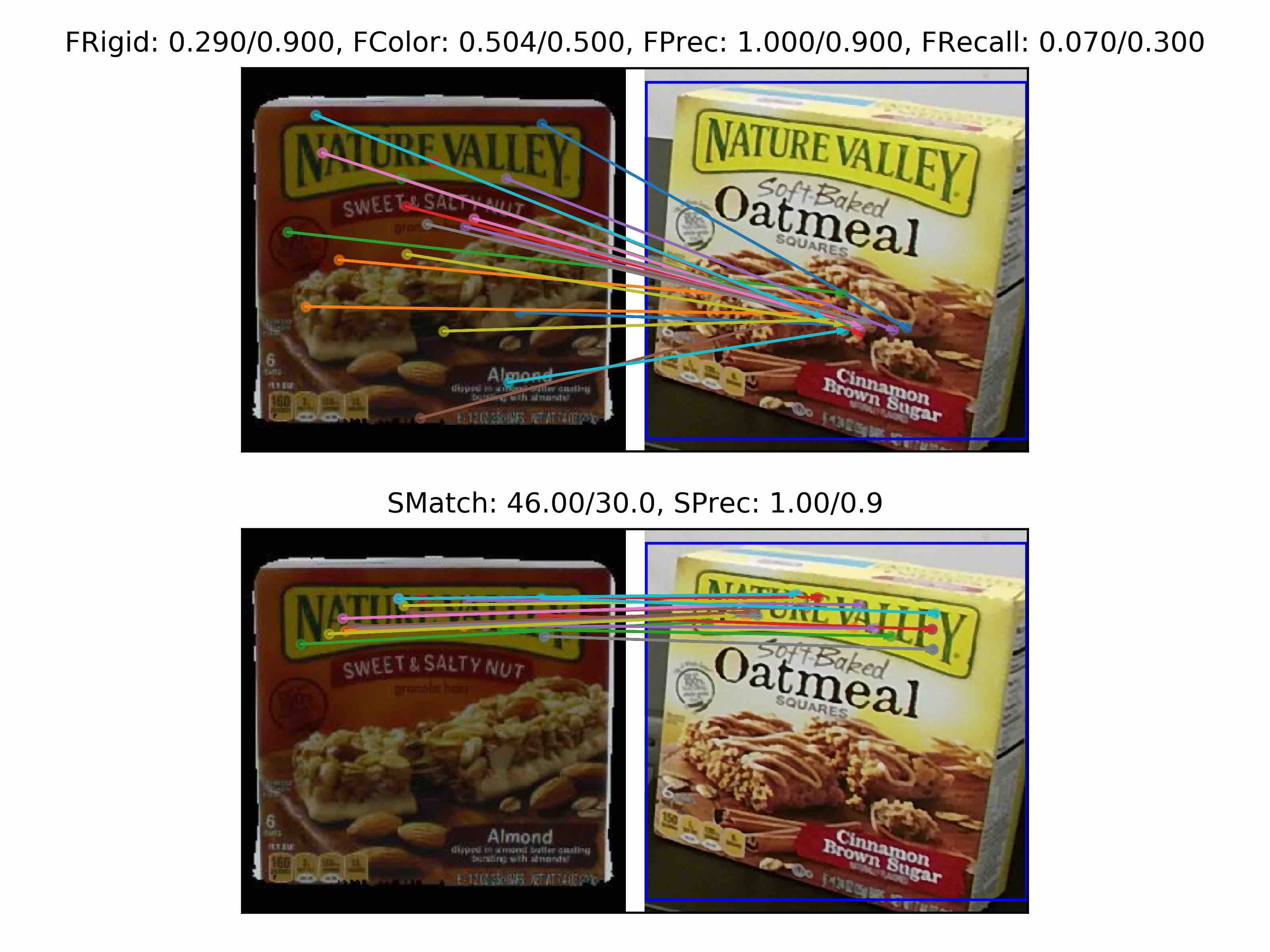}
    \caption{}
    \end{subfigure}
    \begin{subfigure}[t]{\scale\linewidth}
    \includegraphics[width=\textwidth]{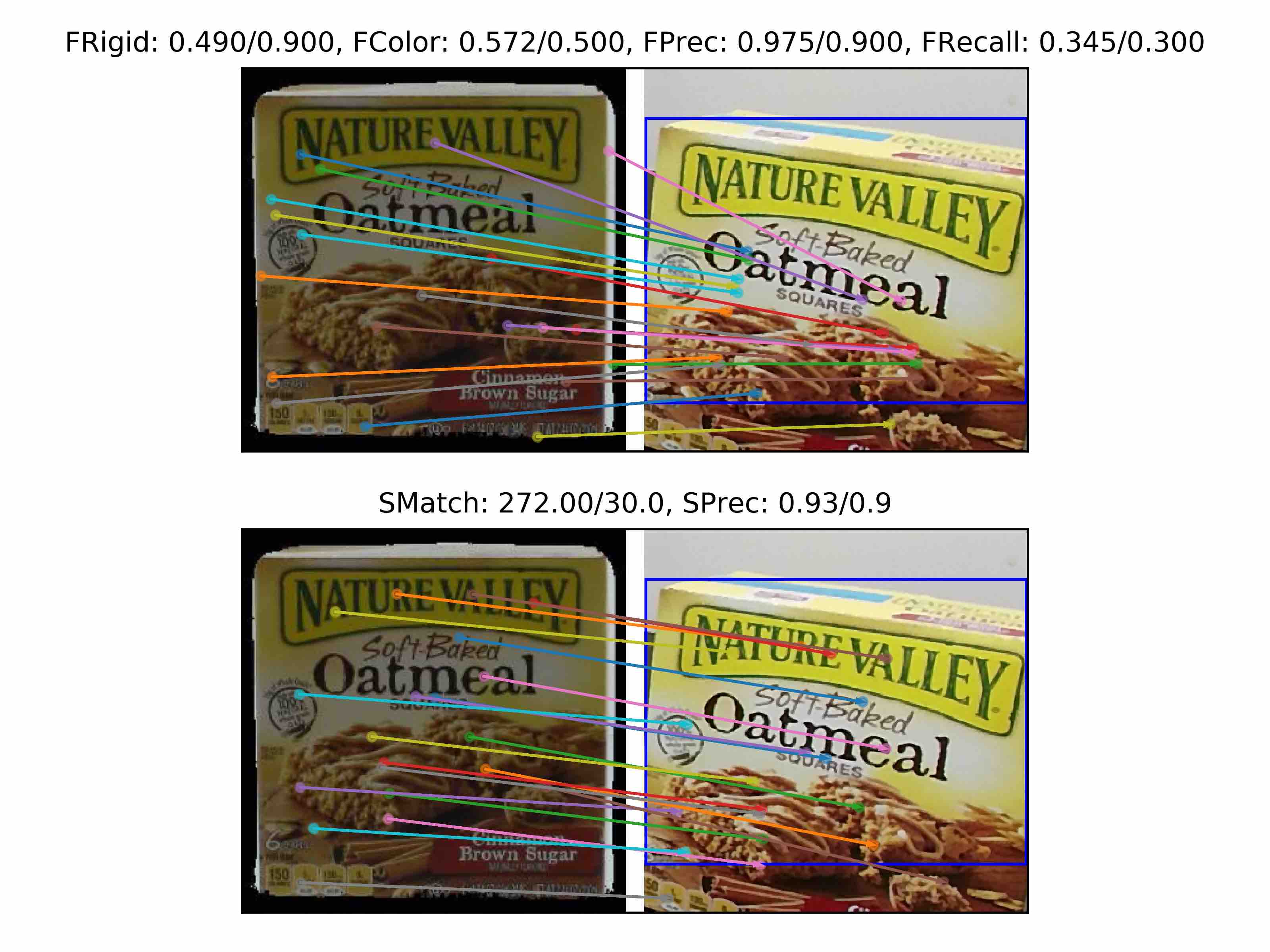}
    \caption{}
    \end{subfigure}
    \\
    \begin{subfigure}[t]{\scale\linewidth}
    \includegraphics[width=\textwidth]{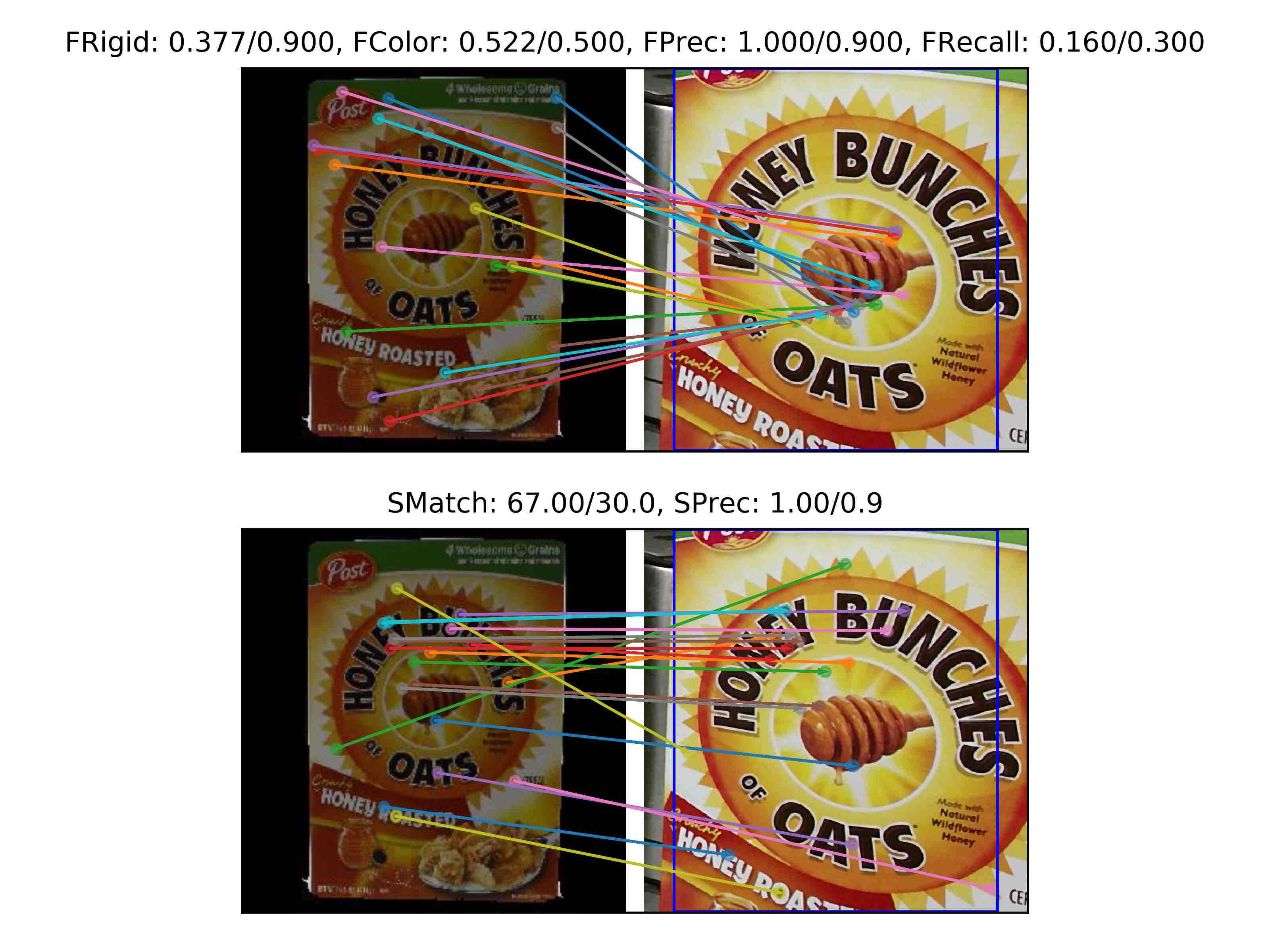}
    \caption{}
    \end{subfigure}
    \begin{subfigure}[t]{\scale\linewidth}
    \includegraphics[width=\textwidth]{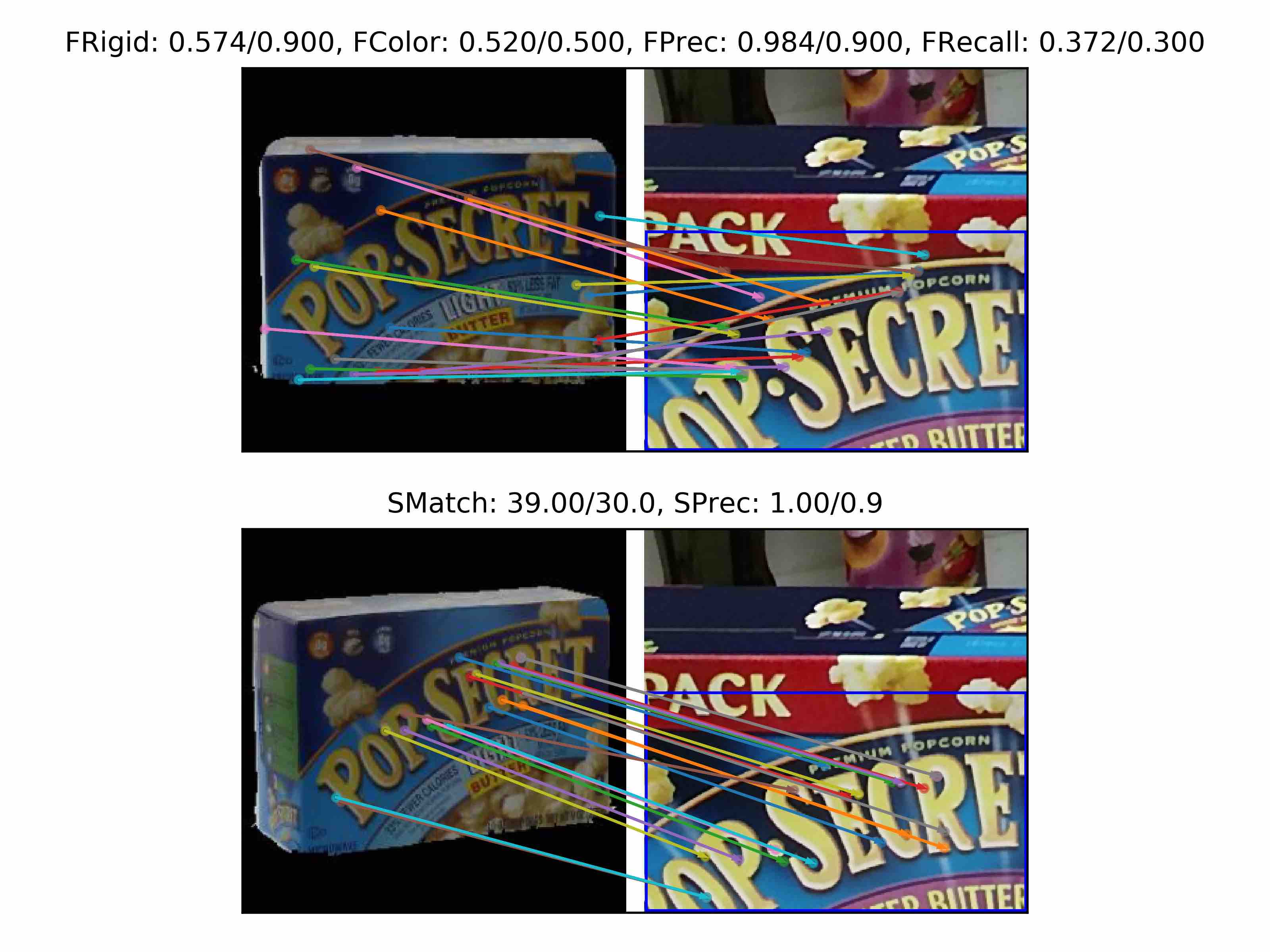}
    \caption{}
    \end{subfigure}
  \caption{False detections which are incorrectly retained by \siftverify. In each image, top: \flowverify, bottom: \siftverify.}
  \label{fig:ffp}
\end{figure*}

\begin{figure*}
  \centering
    \begin{subfigure}[t]{\scale\linewidth}
    \includegraphics[width=\textwidth]{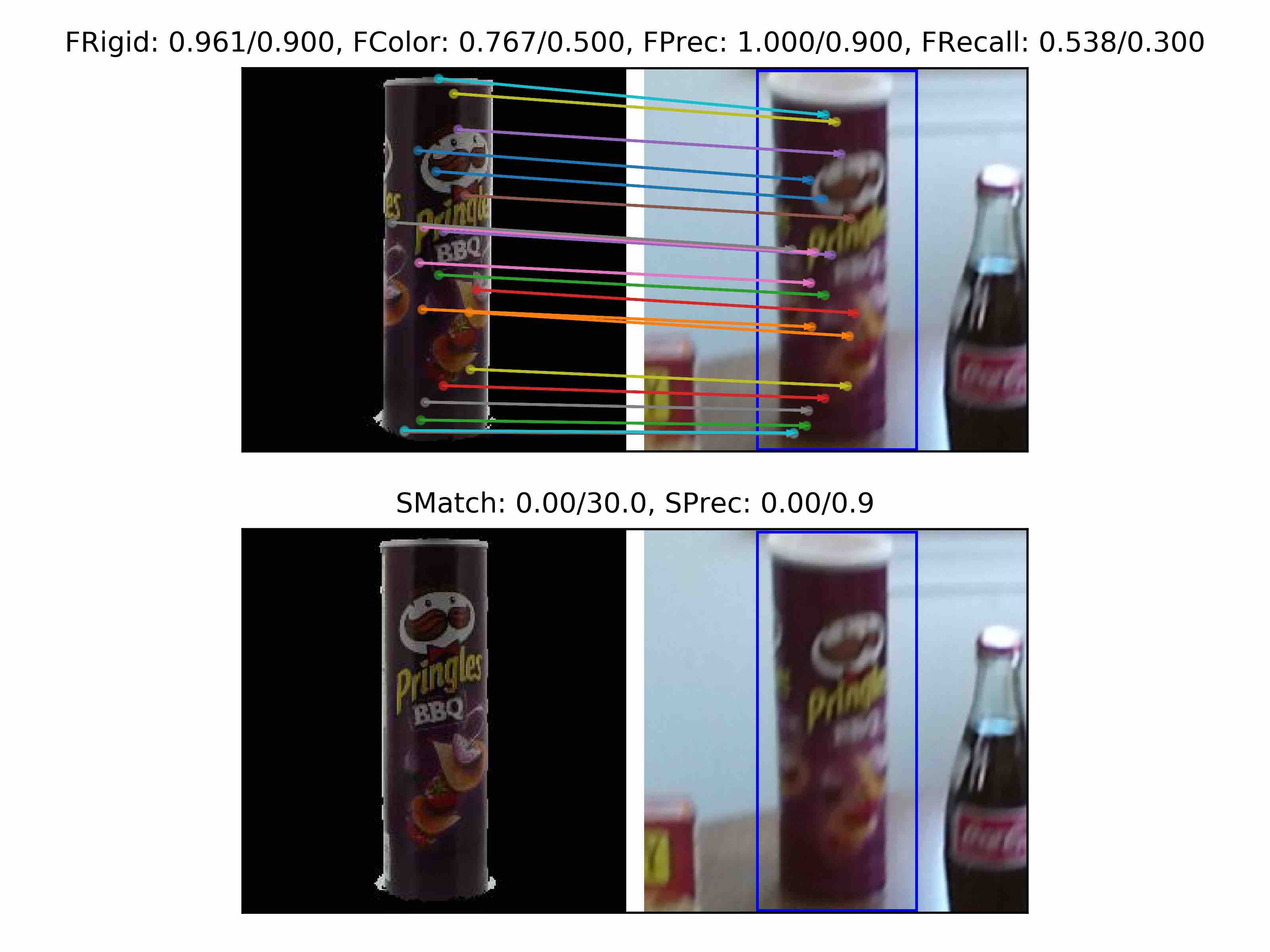}
    \caption{}
    \end{subfigure}
    \begin{subfigure}[t]{\scale\linewidth}
    \includegraphics[width=\textwidth]{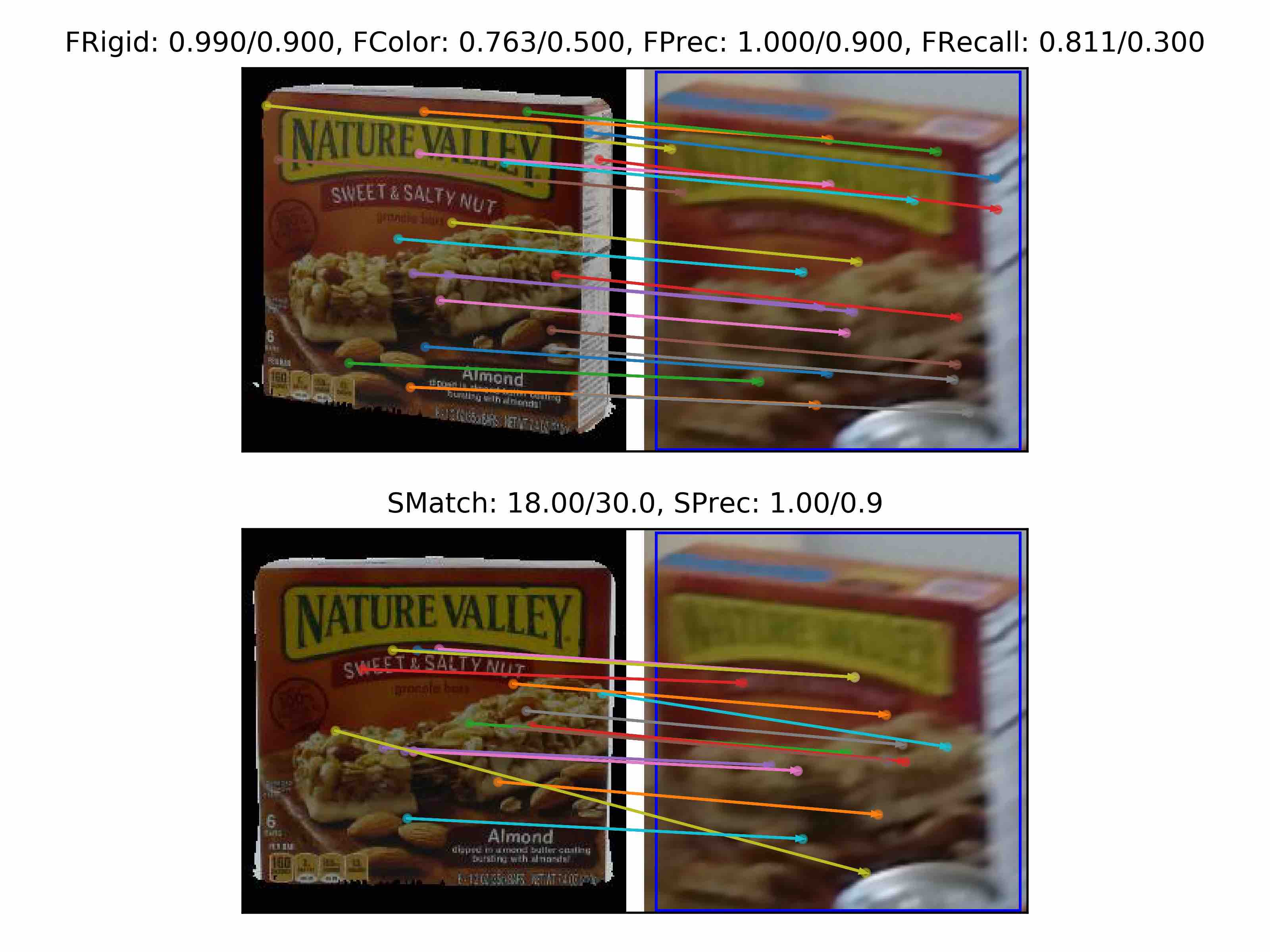}
    \caption{}
    \end{subfigure}\\
    \begin{subfigure}[t]{\scale\linewidth}
    \includegraphics[width=\textwidth]{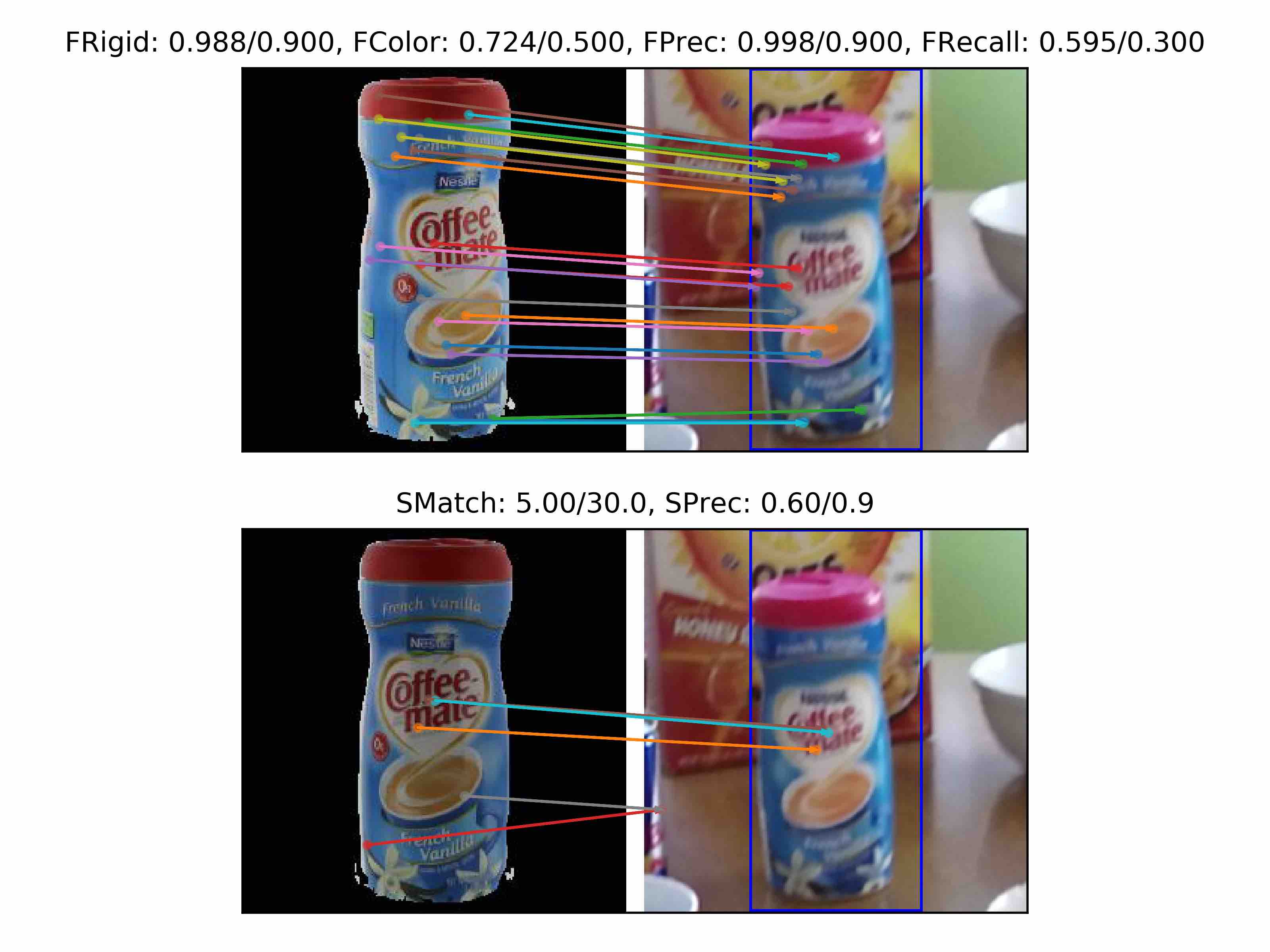}
    \caption{}
    \end{subfigure}
    \begin{subfigure}[t]{\scale\linewidth}
    \includegraphics[width=\textwidth]{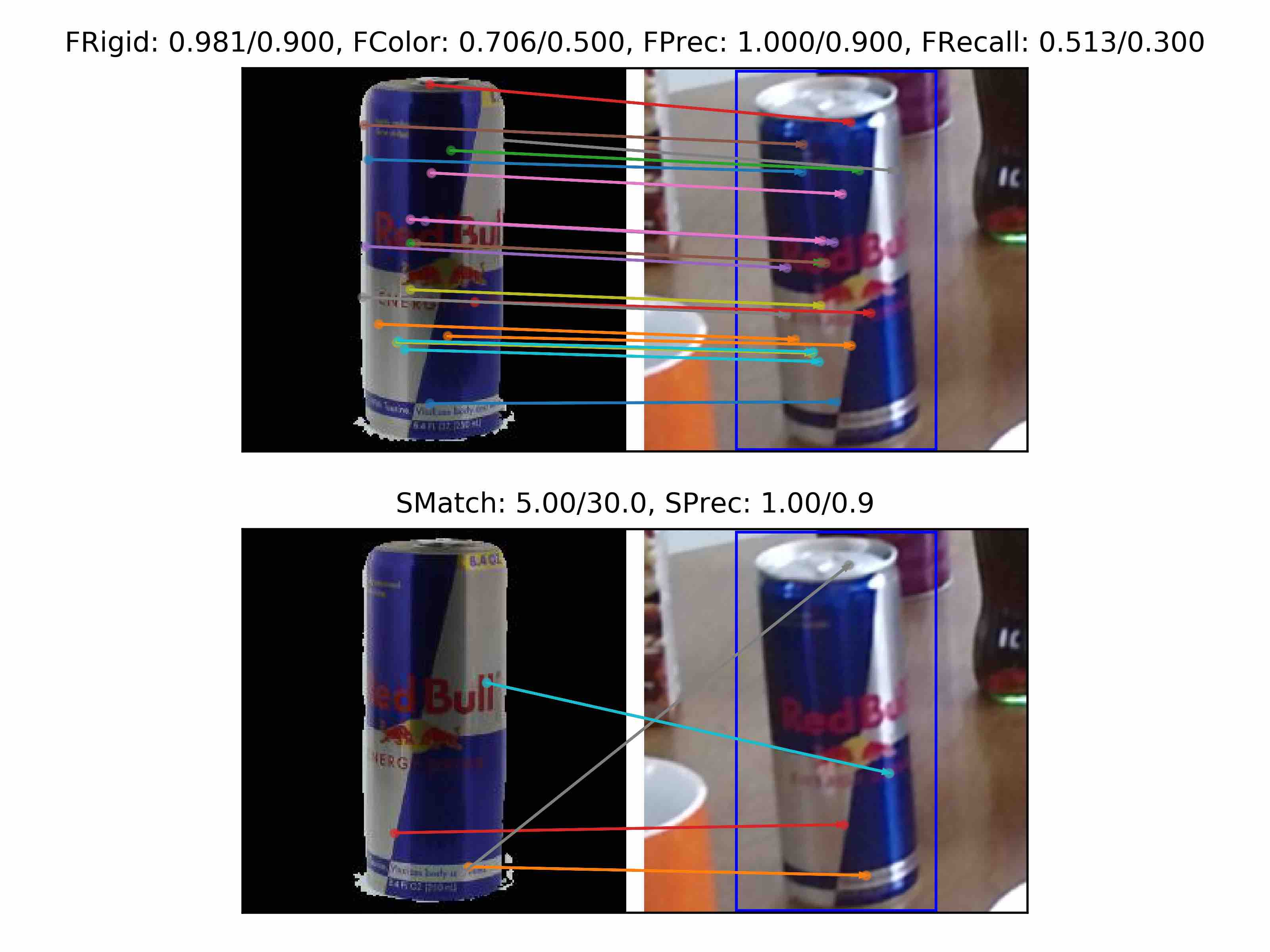}
    \caption{}
    \end{subfigure}
  \caption{True detections that are incorrectly filtered by \siftverify. In each image, top: \flowverify, bottom: \siftverify.}
  \label{fig:ftp}
\end{figure*}

\end{appendices}

\end{document}